%% file: main.tex
\documentclass[10pt,twocolumn,letterpaper]{article}
\usepackage{indentfirst}
\usepackage{tabu}
\usepackage[title]{appendix}
\usepackage{graphicx}
\usepackage{amsmath}
\usepackage{booktabs}
\usepackage{tikz}
\usepackage{rotating}
\usetikzlibrary{arrows.meta,positioning,fit,calc}

\usepackage{cvpr}                

\input{preamble}

\definecolor{cvprblue}{rgb}{0.21,0.49,0.74}
\usepackage[pagebackref,breaklinks,colorlinks,allcolors=cvprblue]{hyperref}

\def\paperID{NTIRE-147} 
\def\confName{CVPR}
\def\confYear{2026}

\title{NTIRE 2026 Challenge on Single Image Reflection Removal in the Wild: Datasets, Results, and Methods}

\author{
Jie Cai\thanks{Jie Cai, Kangning Yang, Zhiyuan Li, Florin-Alexandru Vasluianu, Radu Timofte, Jinlong Li, Jinglin Shen, and Zibo Meng served as the challenge organizers, while all other authors participated as challenge competitors. The team and affiliation information for each author is provided in~\cref{sec:appendix}.}
\and
Kangning Yang\footnotemark[1]
\and
Zhiyuan Li\footnotemark[1]
\and
Florin-Alexandru Vasluianu\footnotemark[1]
\and
Radu Timofte\footnotemark[1]
\and
Jinlong Li\footnotemark[1]
\and
Jinglin Shen\footnotemark[1]
\and
Zibo Meng\footnotemark[1]
\and
Junyan Cao \and Lu Zhao \and Pengwei Liu \and Yuyi Zhang \and Fengjun Guo
\and
Jiagao Hu \and Zepeng Wang \and Fei Wang \and Daiguo Zhou
\and
Yi'ang Chen \and Honghui Zhu \and Mengru Yang \and Yan Luo \and Kui Jiang \and Jin Guo
\and
Jonghyuk Park \and Jae-Young Sim
\and
Wei Zhou \and Hongyu Huang \and Linfeng Li \and Lindong Kong
\and
Saiprasad Meesiyawar \and Misbha Falak Khanpagadi \and Nikhil Akalwadi \and Ramesh Ashok Tabib \and Uma Mudenagudi
\and
Bilel Benjdira \and Anas M. Ali \and Wadii Boulila
\and
Kosuke Shigematsu \and Hiroto Shirono \and Asuka Shin
\and
Guoyi Xu \and Yaoxin Jiang \and Jiajia Liu \and Yaokun Shi \and Jiachen Tu
\and
Shreeniketh Joshi
\and
Jin-Hui Jiang \and Yu-Fan Lin \and Yu-Jou Hsiao \and Chia-Ming Lee \and Fu-En Yang \and Yu-Chiang Frank Wang \and Chih-Chung Hsu
}

\begin{document}
\maketitle
\input{Sections/0_Abstract}    
\input{Sections/1_Introduction}
\input{Sections/2_NTIRE_2026_Challenge}

\input{Sections/3_Challenge_Results}
\input{Sections/4_Challenge_Methods}
\input{Sections/5_AIGC}

\input{Sections/6_Conclusion}

\section*{Acknowledgments}
This work was partially supported by the Humboldt Foundation. We thank the NTIRE 2026 sponsors: OPPO, Kuaishou, and the University of Wurzburg (Computer Vision Lab).

\clearpage
{
\bibliographystyle{ieeenat_fullname}
\bibliography{main}
}

\appendix
\input{Sections/Appendix-A}

\clearpage
\input{Sections/Appendix-B}


\end{document}

%% file: preamble.tex
\usepackage{multirow}

%% file: Sections/0_Abstract.tex
\begin{abstract}

   In this paper, we review the NTIRE 2026 challenge on single-image reflection removal (SIRR) in the wild. SIRR is a fundamental task in image restoration. Despite progress in academic research, most methods are tested on synthetic images or limited real-world images, creating a gap in real-world applications. In this challenge, we provide participants with the OpenRR-5k dataset. This dataset requires participants to process real-world images covering a range of reflection scenarios and intensities, aiming to generate clean images without reflections. The challenge attracted more than 100 registrations, with eleven of them participating in the final testing phase. The top-ranked methods advanced the state-of-the-art reflection removal performance and earned unanimous recognition from five experts in the field. The proposed OpenRR-5k dataset is available at \url{https://huggingface.co/datasets/qiuzhangTiTi/OpenRR-5k}, and the homepage of this challenge is at \url{https://github.com/caijie0620/OpenRR-5k}.
   
\end{abstract}

%% file: Sections/1_Introduction.tex
\section{Introduction}
\label{sec:intro}

SIRR is a critical task in image restoration, focusing on recovering the transmission layer $T$ from an input image $I$ with reflection contamination $R$ caused by different reflective surfaces (e.g., transparent glass).

Over the years, various techniques have been proposed to address the SIRR problem. Traditional methods typically rely on non-learning paradigms to mitigate the ill-posed nature of this problem~\cite{wan2016depth, yang2019fast}. However, these methods typically rely on hand-crafted priors to guide the recovery process, which limits their ability to generalize well to diverse real-world scenarios. To address this issue, deep learning-based methods have been used to model the uncertainty of transmission estimation. Several of the most recent works~\cite{yang2025survey} are summarized in~\cref{tab:1}.

\input{Tables/table_1}

Despite deep learning's progress, the scarcity of high-quality data remains a bottleneck. Real-world dataset construction is often hindered by intensive labor costs and the technical difficulty of achieving precise pixel-level alignment in complex environments. To address this, we propose a novel data collection protocol specifically designed to capture high-quality, aligned image pairs. Based on this protocol, we have collected real-world, diverse, and pixel-aligned datasets: OpenRR-1k~\cite{yang2025openrr} and OpenRR-5k~\cite{cai2025openrr}. These high-quality datasets aim to advance research in reflection removal. OpenRR-1k~\cite{yang2025openrr} served as the benchmark dataset for the NTIRE 2025 SIRR challenge~\cite{yang2025ntire}.

This challenge aims to provide a platform for evaluating models in real-world scenarios, thereby narrowing the gap between academic research and practical photography.

%% file: Tables/table_1.tex
\begin{table*}
\centering   
\caption{\textbf{I}, \textbf{R}, \textbf{T}, and \textbf{E} represent the \textbf{I}nput, \textbf{R}eflection, \textbf{T}ransmission, and \textbf{E}dge map, respectively. The subscripts of \textbf{T} and \textbf{R} represent intermediate process outputs. The Absorption Effect $e$ is introduced in~\cite{zheng2021single} to describe light attenuation as it passes through the glass. The output $residue$ term, proposed in~\cite{hu2023single}, is used to correct errors in the additive reconstruction of the reflection and transmission layers. Language descriptions in~\cite{zhong2024language} provide contextual information about the image layers, assisting in addressing the ill-posed nature of the reflection separation problem.}
\label{tab:1}
\resizebox{0.95\textwidth}{!}{
\begin{tabu}{lllll} 
\toprule
 & \shortstack[l]{\space \\ \space \\ \textbf{Methods} \\ \space } & 
   \shortstack[l]{\space \\ \space \\ \textbf{Venue} \\ \space } & 
   \shortstack[l]{\space \\ \space \\ \textbf{Scheme} \\ \space } & 
   \shortstack[l]{\space \\ \space \\ \textbf{Cross-stage fusion} \\ \space} \\
\toprule
\multirow{4}{*}{Single-stage} & Zhang et al.~\cite{zhang2018single} & CVPR 2018 & $I \rightarrow [T, R]$ & N/A \\ 
\cline{2-5}
                & ERRNet~\cite{wei2019single} & CVPR 2019 & $I \rightarrow T$ & N/A \\ 
\cline{2-5}
                & RobustSIRR~\cite{song2023robust} & CVPR 2023 & $I_{multiscale} \rightarrow T$ & N/A \\ 
\cline{2-5}
                & YTMT~\cite{hu2021trash} & NeurIPS 2021 & $I \rightarrow [T, R]$ & N/A \\ 
\cline{2-5}
                & F2T2-HiT~\cite{cai2025f2t2} & ICIP 2025 & $I \rightarrow T$ & N/A \\
\hline
\multirow{10}{*}{Two-stage}   & CoRRN~\cite{wan2019corrn} & TPAMI 2019 & \begin{tabular}[c]{@{}l@{}}$I \rightarrow E_{T}$ \\ $[I, E_{T}] \rightarrow T$\end{tabular} & Convolutional Fusion \\ 
\cline{2-5}
                & DMGN~\cite{feng2021deep} & TIP 2021 & \begin{tabular}[c]{@{}l@{}}$I \rightarrow [T_{1}, R]$ \\ $[I, T_{1}, R] \rightarrow T$\end{tabular} & Convolutional Fusion \\ 
\cline{2-5}
                & RAGNet~\cite{li2023two} & Appl. Intell. 2023 & \begin{tabular}[c]{@{}l@{}}$I \rightarrow R$ \\$[I, R] \rightarrow T$\end{tabular} & Convolutional Fusion \\ 
\cline{2-5}
                & CEILNet~\cite{fan2017generic} & ICCV 2017 & \begin{tabular}[c]{@{}l@{}}$[I, E_{I}] \rightarrow E_{T}$ \\ $[I, E_{T}] \rightarrow T$\end{tabular} & Concat \\ 
\cline{2-5}
                & DSRNet~\cite{hu2023single} & ICCV 2023 & \begin{tabular}[c]{@{}l@{}}$I \rightarrow (T_{1}, R_{1})$ \\ $(R_{1}, T_{1}) \rightarrow (R, T, residue)$\end{tabular} & N/A \\ 
\cline{2-5}
                & SP-net BT-net~\cite{kim2020single} & CVPR 2020 & \begin{tabular}[c]{@{}l@{}}$I \rightarrow [T_{1}, R_{1}]$ \\ $R_{1} \rightarrow R$\end{tabular} & N/A \\ 
\cline{2-5}
                & Wan et al.~\cite{wan2020reflection} & CVPR 2020 & \begin{tabular}[c]{@{}l@{}}$[I, E_{I}] \rightarrow R_{1}$ \\ $R_{1} \rightarrow R$\end{tabular} & N/A \\ 
\cline{2-5}
                & Zheng et al.~\cite{zheng2021single} & CVPR 2021 & \begin{tabular}[c]{@{}l@{}}$I \rightarrow e$ \\ $[I, e] \rightarrow T$\end{tabular} & Concat \\ 
\cline{2-5}
                & Zhu et al.~\cite{zhu2024revisiting} & CVPR 2024 & \begin{tabular}[c]{@{}l@{}}$I \rightarrow E_{R}$ \\$[I, E_{R}] \rightarrow T$\end{tabular} & Concat \\ 
\cline{2-5}
                & Language-Guided~\cite{zhong2024language} & CVPR 2024 & \begin{tabular}[c]{@{}l@{}}$[I, Texts] \rightarrow R\ or\ T$ \\ $[I, R\ or\ T] \rightarrow T\ or\ R$\end{tabular} & Feature-Level Concat \\ 
\cline{2-5}
                & Cai et al.~\cite{cai2025degradation} & MIPR 2025 & \begin{tabular}[c]{@{}l@{}}$I \rightarrow T_{1}$ \\$T_{1} \rightarrow T$\end{tabular} & N/A \\ 
\cline{2-5}
\hline
\multirow{5}{*}{Multi-stage}  & BDN~\cite{yang2018seeing} & ECCV 2018 & \begin{tabular}[c]{@{}l@{}}$I \rightarrow T_{1}$ \\ $[I, T_{1}] \rightarrow R$ \\ $[I, R] \rightarrow T$\end{tabular} & Concat \\ 
\cline{2-5}
                 & IBCLN~\cite{li2020single} & CVPR 2020 & \begin{tabular}[c]{@{}l@{}}$[I, R_{0}, T_{0}] \rightarrow [R_{1}, T_{1}]$ \\ $[I, R_{1}, T_{1}] \rightarrow [R_{2}, T_{2}]$ \\\end{tabular} & \begin{tabular}[c]{@{}l@{}}Concat \\ Recurrent\end{tabular} \\ 
\cline{2-5}
                 & Chang et al.~\cite{chang2021single} & WACV 2021 & \begin{tabular}[c]{@{}l@{}}$I \rightarrow E_{T}$ \\ $[I, E_{T}] \rightarrow T_{1} \rightarrow R_{1} \rightarrow T_{2}$ \\ $[I, E_{T}, T_{2}] \rightarrow R \rightarrow T$\end{tabular} & \begin{tabular}[c]{@{}l@{}}Concat \\ Recurrent\end{tabular} \\ 
\cline{2-5}
                 & LANet~\cite{dong2021location} & ICCV 2021 & \begin{tabular}[c]{@{}l@{}}$[I, T_{0}] \rightarrow R_{1} \rightarrow T_{1}$ \\ $[I, T_{1}] \rightarrow R_{2} \rightarrow T_{2}$ \\\end{tabular}                                        & \begin{tabular}[c]{@{}l@{}}Concat \\ Recurrent\end{tabular} \\ 
\cline{2-5}
                 & V-DESIRR~\cite{prasad2021v} & ICCV 2021 & \begin{tabular}[c]{@{}l@{}}$I_{1} \rightarrow T_{1}$ \\ $[I_{1}, T_{1}, I_{2}] \rightarrow T_{2}$ \\$[I_{n-1}, T_{n-1}, I_{n}] \rightarrow T$\end{tabular}                             & \begin{tabular}[c]{@{}l@{}}Convolutional Fusion \\ Recurrent\end{tabular}     \\
\cline{2-5}
                 & L-DiffER~\cite{hong2024differ} & ECCV 2024                                                       & \begin{tabular}[c]{@{}l@{}}$[I, Texts] \rightarrow [I_t^c, I_t^s]$ \\ ... \\ $[I_t^c, I_t^s] \rightarrow T$\end{tabular} & Diffusion Model \\
\cline{2-5}
                 & RDNet~\cite{zhao2025reversible} & CVPR 2025 & \begin{tabular}[c]{@{}l@{}}$I \rightarrow [R_{1}, T_{1}]$ \\ $[I, R_{1}, T_{1}] \rightarrow [R_{2}, T_{2}]$ \\ ... \\\end{tabular}                                            & \begin{tabular}[c]{@{}l@{}}Convolutional Fusion \\ Recurrent\end{tabular} \\ 
\bottomrule
\end{tabu}
}
\end{table*}

%% file: Sections/2_NTIRE_2026_Challenge.tex
\begin{figure*}[ht]
    \centering
    \footnotesize
    \includegraphics[width=1.0\textwidth]{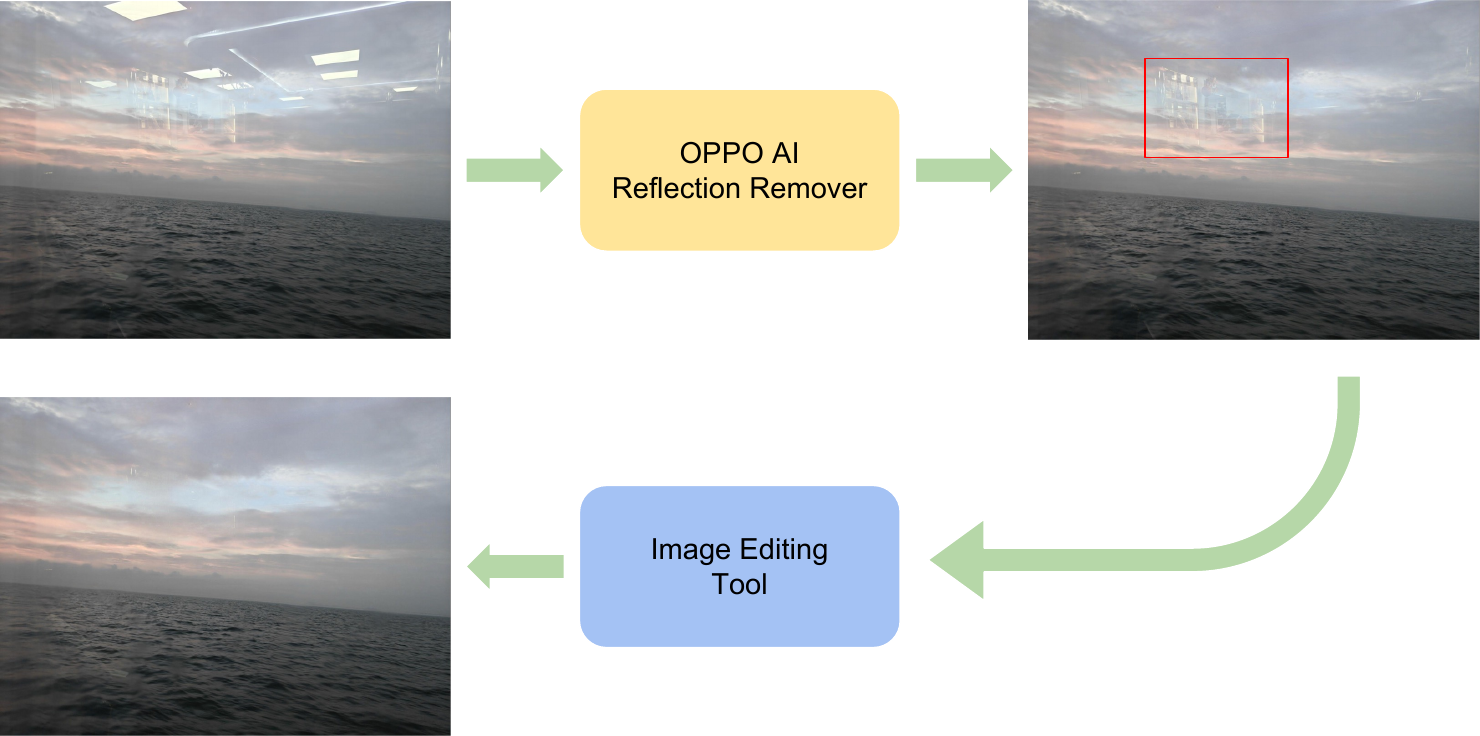}
    \caption{Visualization of paired data generation pipeline for reflection removal.}
    \label{fig:protocol}
\end{figure*}

\section{NTIRE 2026 SIRR Challenge}
\label{sec:challenge}

\subsection{Overview}

This challenge is one of the challenges associated with the NTIRE 2026 Workshop~\footnote{\url{https://www.cvlai.net/ntire/2026/}} on:
deepfake detection~\cite{ntire26deepfake}, 
high-resolution depth~\cite{ntire26hrdepth},
multi-exposure image fusion~\cite{ntire26raim_fusion}, 
AI flash portrait~\cite{ntire26raim_portrait}, 
professional image quality assessment~\cite{ntire26raim_piqa},
light field super-resolution~\cite{ntire26lightsr},
3D content super-resolution~\cite{ntire263dsr},
bitstream-corrupted video restoration~\cite{ntire26videores},
X-AIGC quality assessment~\cite{ntire26XAIGCqa},
shadow removal~\cite{ntire26shadow},
ambient lighting normalization~\cite{ntire26lightnorm},
controllable Bokeh rendering~\cite{ntire26bokeh},
rip current detection and segmentation~\cite{ntire26ripdetseg},
low light image enhancement~\cite{ntire26llie},
high FPS video frame interpolation~\cite{ntire26highfps},
Night-time dehazing~\cite{ntire26nthaze,ntire26nthaze_rep},
learned ISP with unpaired data~\cite{ntire26isp},
short-form UGC video restoration~\cite{ntire26ugcvideo},
raindrop removal for dual-focused images~\cite{ntire26dual_focus},
image super-resolution (x4)~\cite{ntire26srx4},
photography retouching transfer~\cite{ntire26retouching},
mobile real-word super-resolution~\cite{ntire26rwsr},
remote sensing infrared super-resolution~\cite{ntire26rsirsr},
AI-Generated image detection~\cite{ntire26aigendet},
cross-domain few-shot object detection~\cite{ntire26cdfsod},
financial receipt restoration and reasoning~\cite{ntire26finrec},
real-world face restoration~\cite{ntire26faceres},
reflection removal~\cite{ntire26reflection},
anomaly detection of face enhancement~\cite{ntire26anomalydet},
video saliency prediction~\cite{ntire26videosal},
efficient super-resolution~\cite{ntire26effsr},
3d restoration and reconstruction in adverse conditions~\cite{ntire26realx3d},
image denoising~\cite{ntire26denoising},
blind computational aberration correction~\cite{ntire26aberration},
event-based image deblurring~\cite{ntire26eventblurr},
efficient burst HDR and restoration~\cite{ntire26bursthdr},
low-light enhancement: `twilight cowboy'~\cite{ntire26twilight},
and efficient low light image enhancement~\cite{ntire26effllie}.

\par\vspace{6pt}
The objectives of this SIRR challenge are as follows:

\begin{itemize}
    \item To provide the community with a large-scale real-world SIRR benchmark encompassing diverse scenarios and objective performance protocols.
    \item To stimulate the exploration of SIRR architectures capable of handling the inherent complexity and intensity variations of real-world reflections.
    \item To bridge the existing trajectory between laboratory-scale research and robust consumer-grade imaging solutions.
\end{itemize}

\subsection{OpenRR-5k Dataset}
\label{subsec:dataset}
In many practical scenarios, the acquisition of precisely aligned ground-truth images is very difficult. Researchers typically rely on the utilization of props such as glass and cloth in their methodologies. After capturing the blended images, they construct reflection pairs (e.g., $(I,T)$, $(I, I - R)$, etc.) by either removing the glass or blocking the background or reflection light with light-absorbing black velvet cloth~\cite{li2020single, lei2020polarized, lei2022categorized, zhu2024revisiting}. For example, ~\citet{li2020single} obtained transmitted images by manually removing the glass. More recently, ~\citet{zhu2024revisiting} proposed a new data collection pipeline that involves blocking all reflection lights generated by the surrounding environment. However, environmental factors like wind and vibration can induce misalignment and color shifts, degrading the quality of reflection pairs.

\begin{figure*}[ht]
    \centering
    \footnotesize
    \includegraphics[width=1.0\textwidth]{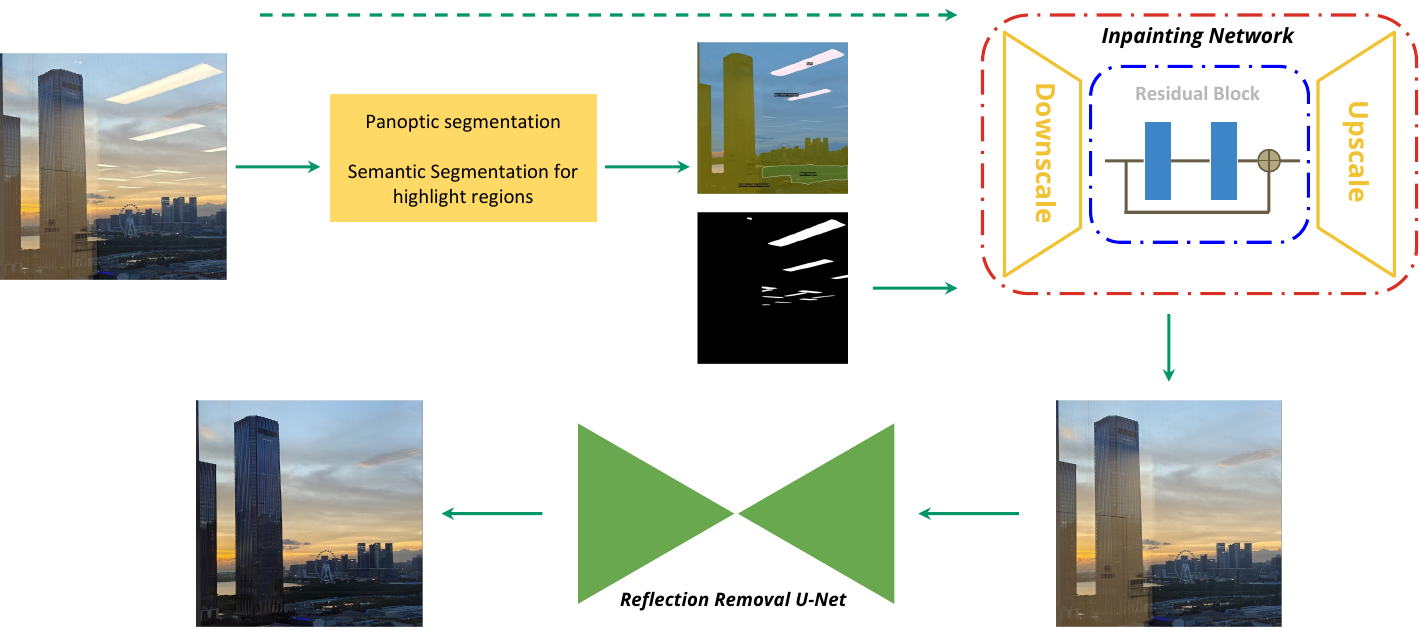}
    \caption{Visualization of the OPPO AI Reflection Remover pipeline deployed on Find X8 Ultra, using the April 2025 algorithm version. The OPPO reflection removal method adopts a two-stage framework: (1) In the first stage, the system utilizes YOLO and YOSO for robust detection of strong reflection masks, followed by LaMa-based inpainting to fill the detected regions. (2) In the second stage, NAFNet is employed to perform fine-grained reflection removal, further restoring the underlying transmission layer and ensuring high visual fidelity.}
    \label{fig:oppo_method}
\end{figure*}

In this challenge, we propose a novel data collection protocol specifically designed to capture high-quality pairs of transmission and blended images. As illustrated in~\cref{fig:protocol}, our proposed data collection protocol consists of two main phases. In the first phase, we leveraged advanced AI-driven tools to generate ground-truth images rather than traditional manual methods such as removing glass or using light-absorbing black cloth. In this challenge, we adopted the AI reflection removal editor integrated within the OPPO smartphone~\cite{cai2025degradation}\footnote{\href{https://www.oppo.com/en/newsroom/stories/coloros-15-launch-ai/}{https://www.oppo.com/en/newsroom/stories/coloros-15-launch-ai/}} to obtain the initial transmission results since it is one of the few effective and widely used tools currently available on the market. 
As shown in Fig.~\ref{fig:oppo_method}, the OPPO proposed architecture performs reflection artifact removal by first using YOLO/YOSO for reflection detection, followed by LaMa~\cite{suvorov2022resolution} for inpainting the masked regions, and finally applying NAFNet~\cite{chen2022nafnet} for reflection removal.
To ensure the superior quality of the ground-truth transmission images, we implemented a post-processing refinement procedure to mitigate any subtle residual reflections or artifacts. By leveraging professional image editing suites such as Adobe Photoshop and MeituPic for meticulous manual adjustments, we produced a final dataset characterized by high visual fidelity, making it highly suitable for both model training and performance evaluation.

\begin{figure*}[ht]
    \centering
    \footnotesize
    \includegraphics[width=1.0\textwidth]{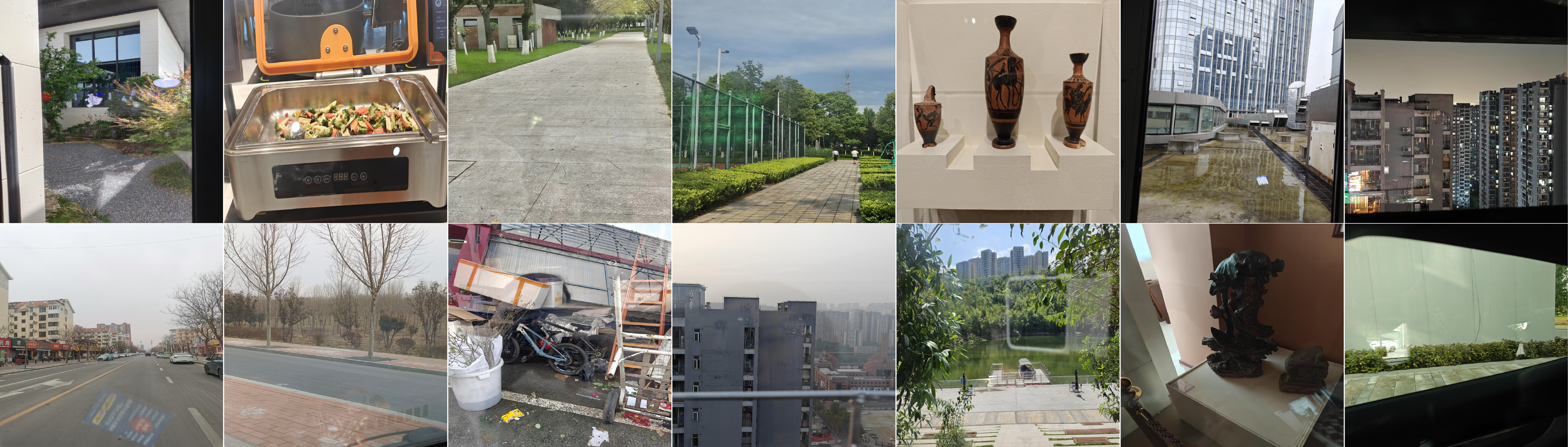}
    \caption{Samples of the OpenRR-5k dataset.}
    \label{fig:example}	 
\end{figure*}

\begin{figure}[ht]
    \centering
    \footnotesize
    \includegraphics[width=0.5\textwidth]{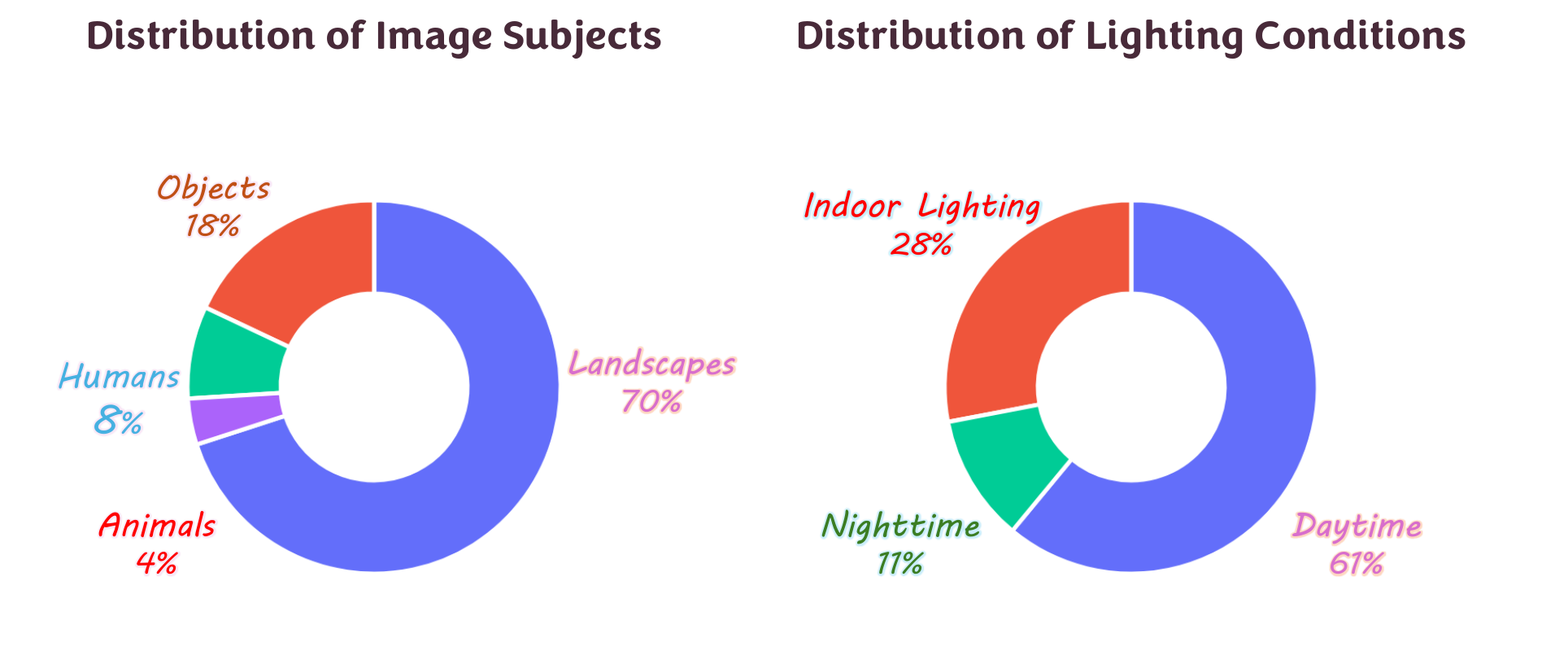}
    \caption{The category distribution of our OpenRR-5k$_{test}$ dataset.}
    \label{fig:statistics}	 
\end{figure}

Following this protocol, we collected a total of 5,300 high-quality pairs of real-world images for this challenge and constructed our OpenRR-5k dataset~\cite{cai2025openrr} as shown in~\cref{fig:example}. We selected 5,000 image pairs for the training set (OpenRR-5k$_{train}$) and 300 image pairs for the validation set (OpenRR-5k$_{val}$). We also collected 100 input-only images for the test set (OpenRR-5k$_{test}$), specifically reserved for evaluating the performance and generalization of the participants' methods in a blind-test manner.
In addition, \cref{fig:statistics} provides a comprehensive summary of the categorical composition of our OpenRR-5k$_{test}$ dataset from two perspectives: image subjects and lighting conditions. Compared to existing open datasets, our dataset in this challenge not only encompasses more diverse scenarios but also comprises \textbf{TRUE} and \textbf{GENUINE} reflection scenarios directly from real-world environments, without relying on artificial setups or simulated reflections. This helps the community evaluate their models more effectively and gain a deeper understanding of the shortcomings in practical applications.

\subsection{Challenge Phases}
There are two phases in the challenge: (1) development/validation phase and (2) testing phase.

\textbf{Development and Validation Phase}: Participants have access to both training and validation data (see Sec.~\ref{subsec:dataset} for dataset details). The training data comprises 5,000 paired images. Each pair consists of an input blended image $I$ and its corresponding ground-truth transmission image $T$. Similarly, the validation data contain another 300 pairs of data samples, but only the input blended images are provided to the participants. Participants can upload their results to the validation server to calculate PSNR/SSIM/LPIPS metrics and receive feedback. 

The ground-truth references for the validation set will be made publicly available on Hugging Face upon the official conclusion of the entire challenge.

\textbf{Testing Phase}: For the blind testing phase, 100 images lacking ground-truth labels are released to the participants. The final submission requires the inclusion of model outputs for both validation and test sets, accompanied by the corresponding source code and comprehensive fact sheets to ensure reproducibility.

\subsection{Evaluation Metrics}

\subsubsection{Objective Metrics on Validation Set}
\label{subsubsection:quantitative}
To quantitatively evaluate the performance of different models, we adopt several standard metrics, including PSNR, SSIM, LPIPS, DISTS, and NIQE. These assessments are conducted by comparing the model outputs with their corresponding ground-truth references\footnote{The evaluation script is publicly available at: \href{https://github.com/caijie0620/OpenRR-5k/blob/main/evaluate.py}{https://github.com/caijie0620/OpenRR-5k/evaluate.py}}.

\subsubsection{Subjective Evaluation on Test Set}
\label{subsubsection:subjective}
We assessed the perceptual quality of the reflection removal results through visual examination. Specifically, we invited five experienced practitioners and conducted a comprehensive user study. The following criteria were taken into account during the evaluation:

\begin{itemize}
    \item \textbf{Reflection Removal Cleanliness $(C)$}: Both strong and weak reflections should be removed as completely and cleanly as possible, without leaving any residue. 
    \item \textbf{Artifacts $(A)$}: Efforts should be made to minimize perceptual artifacts and ensure that no authentic image content is erroneously removed or unnaturally restored.
    \item \textbf{Overall Image Quality $(Q)$}: The output image should have better image quality (including color fidelity, texture, sharpness, detail preservation, exposure, and contrast) than the input reflection-contaminated image.
\end{itemize}

\textbf{The final score $(S)$} is determined by calculating the weighted average of the three criteria mentioned above. 
\begin{equation}
S = 0.25 C + 0.25 A + 0.5 Q
\end{equation}

\subsection{Final Challenge Ranking}
Participants require to submit a single model checkpoint, which we will use to conduct evaluation on both the validation and test sets.
The challenge ranking was conducted as follows. We collected both objective metrics on the validation set and subjective evaluation scores on the test set for all eleven teams, as shown in Table~\ref{tab:performance_comparison}. However, the final challenge ranking was determined solely based on the subjective scores achieved on the test set.

%% file: Sections/3_Challenge_Results.tex
\section{Challenge Results and Analysis}
\label{sec:results}

\par\vspace{6pt}
\textit{\textbf{NTIRE 2026 SIRR Challenge Award Winners:}}:
\par\vspace{1pt}
\begin{itemize}[leftmargin=*, noitemsep, label={}]
    \item \textit{1st Prize:} RRay
    \item \textit{2nd Prizes:} Xreflect Master, AIIALab
    \item \textit{3rd Prizes:} VIP Lab, YuFans, KLETech-CEVI
\end{itemize}
\par\vspace{6pt}

The challenge garnered significant interest, with over 100 registrations and 1,000 submissions. Ultimately, eleven teams successfully completed the testing phase, providing comprehensive submissions that include model outputs, source code, and technical fact sheets. A summary of the participating methodologies is presented in Table~\ref{tab:method_comparison} and~\cref{sec:methods}, with extensive team profiles detailed in~\cref{sec:appendix}.

\begin{figure*}[ht]
    \centering
    \includegraphics[width=1.0\linewidth]{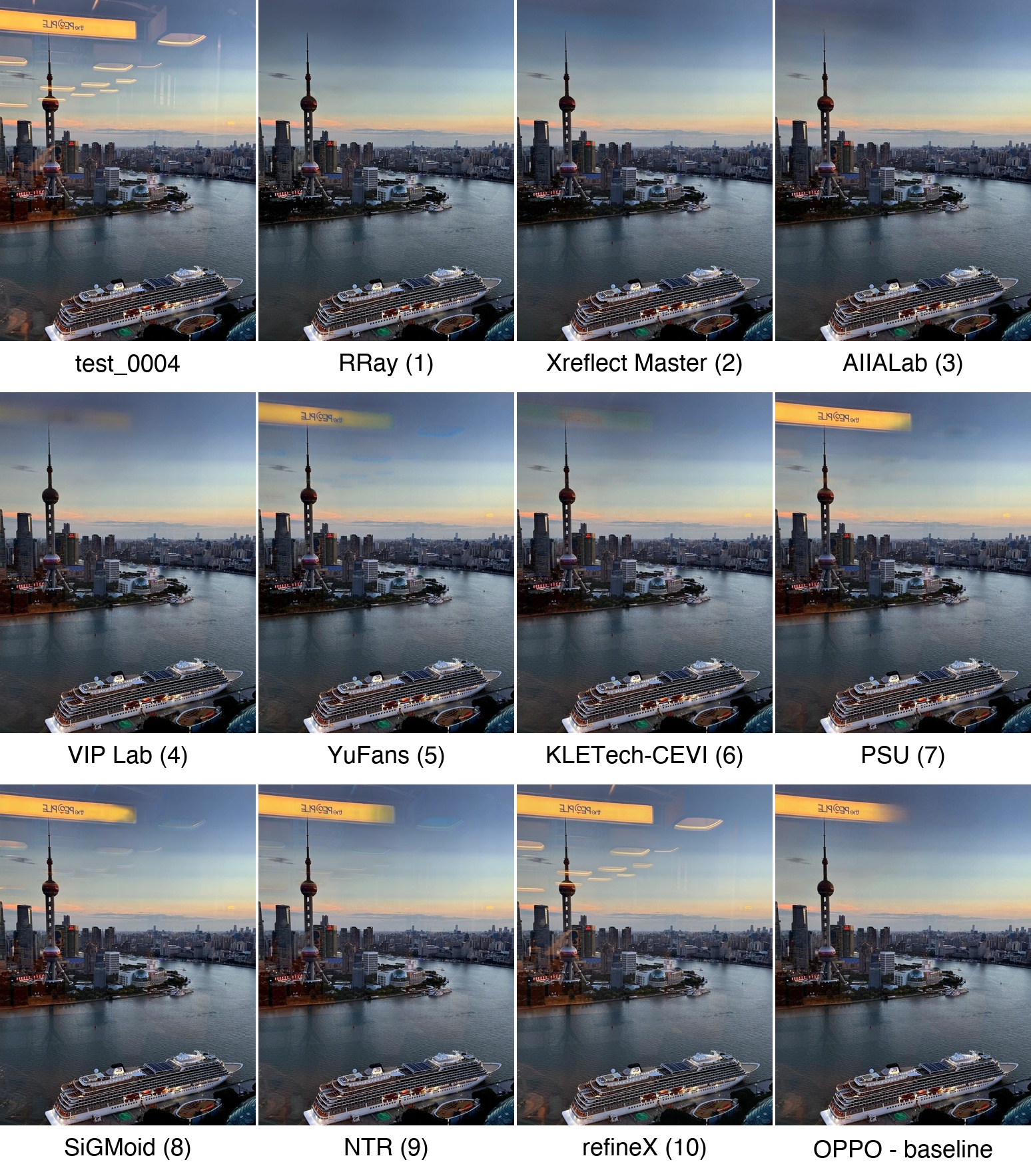}
    \caption{Visual comparison on $test_{0004}$ shows that top methods effectively remove reflections while preserving details, whereas lower-ranked methods suffer from residual reflections, over-smoothing, or color distortions, consistent with their lower subjective scores. For instance, while methods from VIP Lab, YuFans, and KLETech-CEVI are capable of suppressing intense reflections, they struggle to faithfully recover the underlying background details.}
    \label{fig:Visualization-1}
\end{figure*}

\begin{figure*}[ht]
    \centering
    \includegraphics[width=1.0\linewidth]{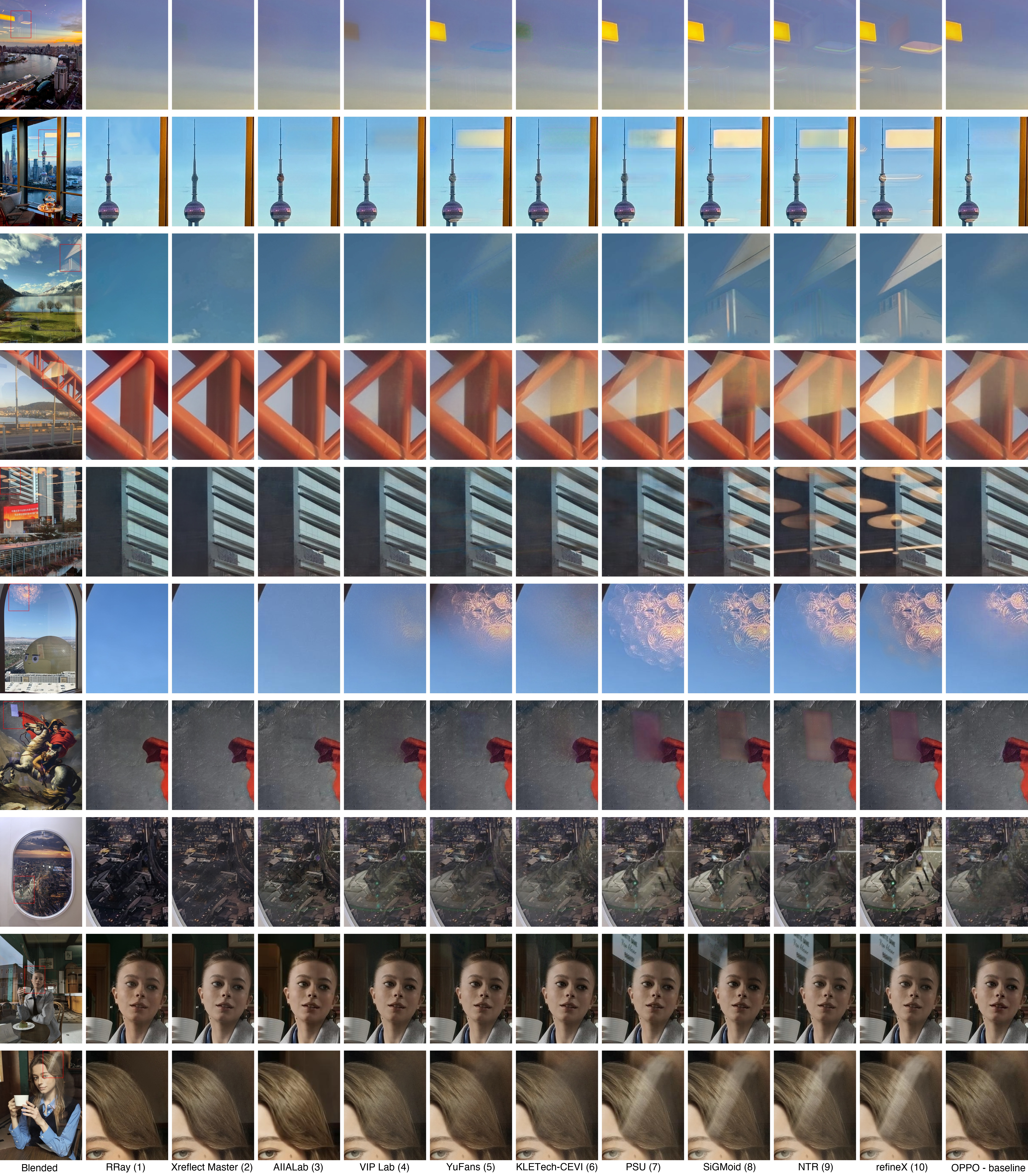}
    \caption{Visual comparison across diverse OpenRR-5k$_{test}$ real-world scenes. Top-ranked methods (RRay, Xreflect Master, and AIIALab) consistently achieve superior reflection removal while preserving fine details and faithful colors. In contrast, lower-ranked methods often exhibit residual reflections, over-smoothing, or color distortions, which aligns with their lower subjective rankings. The ten rows from top to bottom correspond to test samples $test_{0005}$, $test_{0006}$, $test_{0023}$, $test_{0026}$, $test_{0029}$, $test_{0049}$, $test_{0075}$, $test_{0077}$, $test_{0096}$, and $test_{0099}$, respectively.}
    \label{fig:Visualization-2}
\end{figure*}

\input{Tables/table_2}

The final evaluation protocol integrates both objective metrics and subjective human assessments for all eleven participating teams, as summarized in Table~\ref{tab:performance_comparison}. Notably, the official challenge ranking is determined solely based on the subjective scores obtained on the test set. Representative visual comparisons are presented in~\cref{fig:Visualization-1} and~\cref{fig:Visualization-2}, with additional qualitative results provided in~\cref{sec:appendix-b}.

Several noteworthy observations can be drawn from the results. Team RRay achieves the highest subjective score (4.45), while also demonstrating strong performance across objective metrics, including the best PSNR (36.1688) and competitive SSIM (0.9758), LPIPS (0.0235), and DISTS (0.0135). Team Xreflect Master ranks second in subjective evaluation (4.31) and exhibits similarly strong objective performance, achieving a PSNR of 36.0496 and the lowest DISTS score (0.0127) among all participants. Team AIIALab, ranked third, also shows consistently competitive results across both objective and subjective evaluations, indicating a well-balanced design.

Beyond these observations, several important insights emerge. While PSNR exhibits a moderate correlation with subjective performance, particularly among top-ranked methods due to the realistic characteristics of the test set, this relationship is not strictly monotonic. Methods with comparable PSNR values (e.g., YuFans and VIP Lab) can still demonstrate noticeable differences in subjective scores, highlighting the limitations of objective metrics in capturing perceptual quality. Furthermore, as shown in Table~\ref{tab:method_comparison}, top-performing approaches consistently adopt multi-stage or refinement-based designs, underscoring the importance of progressive restoration for handling complex reflection patterns. It is also worth noting that leading methods typically leverage additional external training data beyond the OpenRR-5k$_{train}$ dataset, suggesting that large-scale and diverse data remain a key factor for achieving state-of-the-art performance. Overall, the results reveal a clear perception–distortion trade-off, where optimizing pixel-level fidelity alone is insufficient for achieving high perceptual quality in real-world reflection removal.

%% file: Tables/table_2.tex
\begin{table*}[htbp]
\centering
\renewcommand{\arraystretch}{1.5} 
\begin{tabular}{clcccccc}
\hline
\hline
& \multicolumn{5}{c}{\textbf{Validation Set}} & \textbf{Test Set} \\
Team Name & PSNR $\uparrow$ & SSIM $\uparrow$ & LPIPS $\downarrow$ & DISTS $\downarrow$ & NIQE $\downarrow$ & Subjective Score $\uparrow$ \\
\hline
RRay              & \textbf{36.1688$^{(1)}$}  & 0.9758$^{(3)}$  & 0.0235$^{(4)}$  & \underline{0.0135$^{(2)}$}  & \underline{3.7375$^{(2)}$}  & \textbf{4.45$^{(1)}$} \\
Xreflect Master   & \underline{36.0496$^{(2)}$}  & \textbf{0.9776$^{(1)}$}  & \textbf{0.0210$^{(1)}$}  & \textbf{0.0127$^{(1)}$}  & 3.7648$^{(6)}$ & \underline{4.31$^{(2)}$} \\
AIIALab           & 35.3799$^{(3)}$  & 0.9750$^{(4)}$  & \underline{0.0231$^{(2)}$}  & 0.0155$^{(5)}$  & 3.7737$^{(9)}$  & 4.23$^{(3)}$  \\
VIP Lab           & 34.6872$^{(5)}$  & \underline{0.9766$^{(2)}$}  & \underline{0.0231$^{(2)}$}  & 0.0148$^{(3)}$  & \textbf{3.7218$^{(1)}$}  & 3.85$^{(4)}$  \\
YuFans            & 34.9062$^{(4)}$  & 0.9738$^{(8)}$  & 0.0257$^{(6)}$  & 0.0159$^{(6)}$  & 3.7783$^{(10)}$ & 3.59$^{(5)}$  \\
KLETech-CEVI      & 34.5375$^{(6)}$  & 0.9748$^{(6)}$  & 0.0242$^{(5)}$  & 0.0150$^{(4)}$  & 3.7566$^{(4)}$  & 3.57$^{(6)}$  \\
PSU               & 34.5148$^{(7)}$  & 0.9746$^{(7)}$  & 0.0282$^{(7)}$  & 0.0191$^{(7)}$  & 3.7559$^{(3)}$  & 3.25$^{(7)}$  \\
SiGMoid           & 34.2792$^{(8)}$  & 0.9749$^{(5)}$  & 0.0289$^{(8)}$  & 0.0205$^{(8)}$  & 3.7691$^{(8)}$  & 3.09$^{(8)}$  \\
NTR               & 33.9679$^{(9)}$  & 0.9729$^{(9)}$  & 0.0323$^{(9)}$  & 0.0228$^{(9)}$  & 3.7639$^{(5)}$  & 3.01$^{(9)}$  \\
refineX           & 30.5993$^{(10)}$ & 0.9715$^{(10)}$ & 0.0378$^{(10)}$ & 0.0291$^{(10)}$ & 3.7675$^{(7)}$  & 2.55$^{(10)}$  \\
\hline
OPPO - baseline   & 31.6173 & 0.9229 & 0.0566 & 0.0539 & 3.6205 & \footnotesize{baseline} \\
\hline
ACVLAB            & 31.6886 & 0.9241 & 0.0541 & 0.0551 & 3.6393 & \footnotesize{Late sub.} \\
\hline
\hline
\end{tabular}
\caption{Objective and subjective results on the NTIRE 2026 SIRR Challenge. Note: objective metrics are computed on the OpenRR-5k$_{val}$; subjective scores are calculated on the OpenRR-5k$_{test}$. NTIRE 2026 SIRR in the Wild Challenge Award Winners: 1st Prize: RRay; 2nd Prizes: Xreflect Master, AIIALab; 3rd Prizes: VIP Lab, YuFans, KLETech-CEVI.}
\label{tab:performance_comparison}
\end{table*}

%% file: Sections/4_Challenge_Methods.tex
\input{Tables/table_3}

\clearpage

\section{Challenge Methods}
\label{sec:methods}

To facilitate a clear comparison of the diverse approaches, we consolidate the key technical specifications and performance characteristics of all participating methods in Table~\ref{tab:method_comparison}. Subsequently, comprehensive descriptions of the eleven competing methodologies are provided in the following subsections.

\subsection{RRay -- A Two-Stage Cascaded Network for Single Image Reflection Removal in the Wild}

\begin{figure}[ht]
    \centering
    \includegraphics[width=1.0\linewidth]{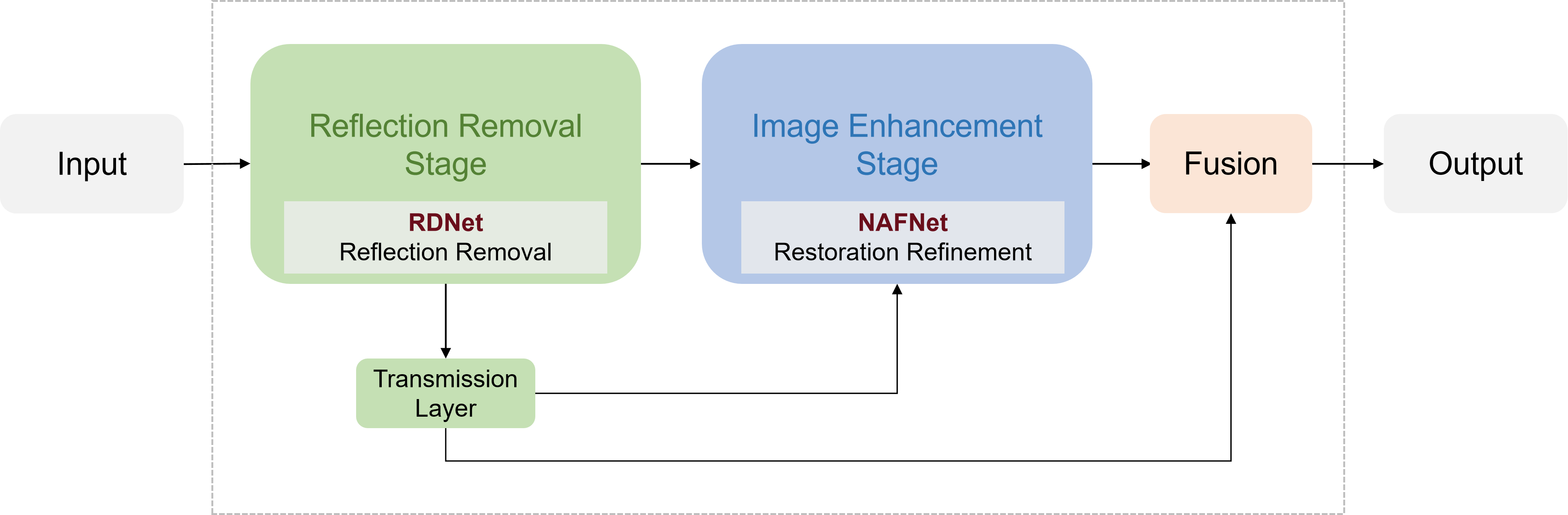}
    \caption{An overview of RdNafNet for SIRR.}
    \label{fig:rray_pipeline}
\end{figure}

As shown in Fig.~\ref{fig:rray_pipeline}, our proposed framework, RdNafNet, is designed as a two-stage cascaded pipeline to address the detail loss issue in single-stage reflection removal. 
Specifically, our empirical observations reveal that while RDNet~\cite{rdnet} can effectively eliminate prominent reflections, it inevitably leads to the loss of fine details in regions originally corrupted by reflections. 
To mitigate this degradation, we decouple the entire restoration process into two consecutive stages. 
The first stage adopts RDNet~\cite{rdnet} as the backbone, targeting the removal of visible reflections from the input image and producing an intermediate result. 
The second stage, by contrast, focuses on enhancing the fine details of this intermediate output, for which we employ NAFNet~\cite{chen2022nafnet} as the backbone to perform more fine-grained image restoration.

Beyond the competition-provided training dataset, we further enrich our training set by supplementing it with both real-world and synthetic datasets. Specifically, for real-world scenarios, we adopt 289 image pairs derived from the ``Real-Nature'' dataset\footnote{\label{fn:xreflection}\url{https://github.com/hainuo-wang/XReflection}}, along with an additional 29,771 real-world paired samples from RRW~\cite{zhu2023Weather}. For the synthetic dataset, we construct representative simulated samples that mimic real reflection degradation patterns, such as ghosting and local reflections. We dynamically synthesize training data using 42,736 images\footref{fn:xreflection} during the training phase.

Implemented based on the PyTorch and BasicSR frameworks, our model is trained with the AdamW optimizer (initial learning rate = 0.0001, halved every 40,000 iterations) on 8 NVIDIA A100 GPUs. Specifically, we first train on 384×384 randomly cropped patches for 100,000 iterations with MSE loss, gradient loss, and VGG perceptual loss. We then increase the patch size to 768×768 and add adversarial loss for an additional 50,000 iterations. For testing, we apply test-time augmentation (TTA) via horizontal, vertical, and combined horizontal/vertical flips of the input image.

Finally, we perform a weighted fusion on the outputs of the model trained with adversarial loss and the original model, enabling an adaptive combination of their complementary strengths for improved reconstruction quality.

\subsection{Xreflect Master -- Scaling RDNet with Diffusion Distillation for Reflection Removal}

\begin{figure}[ht]
    \centering
    \includegraphics[width=1.0\linewidth]{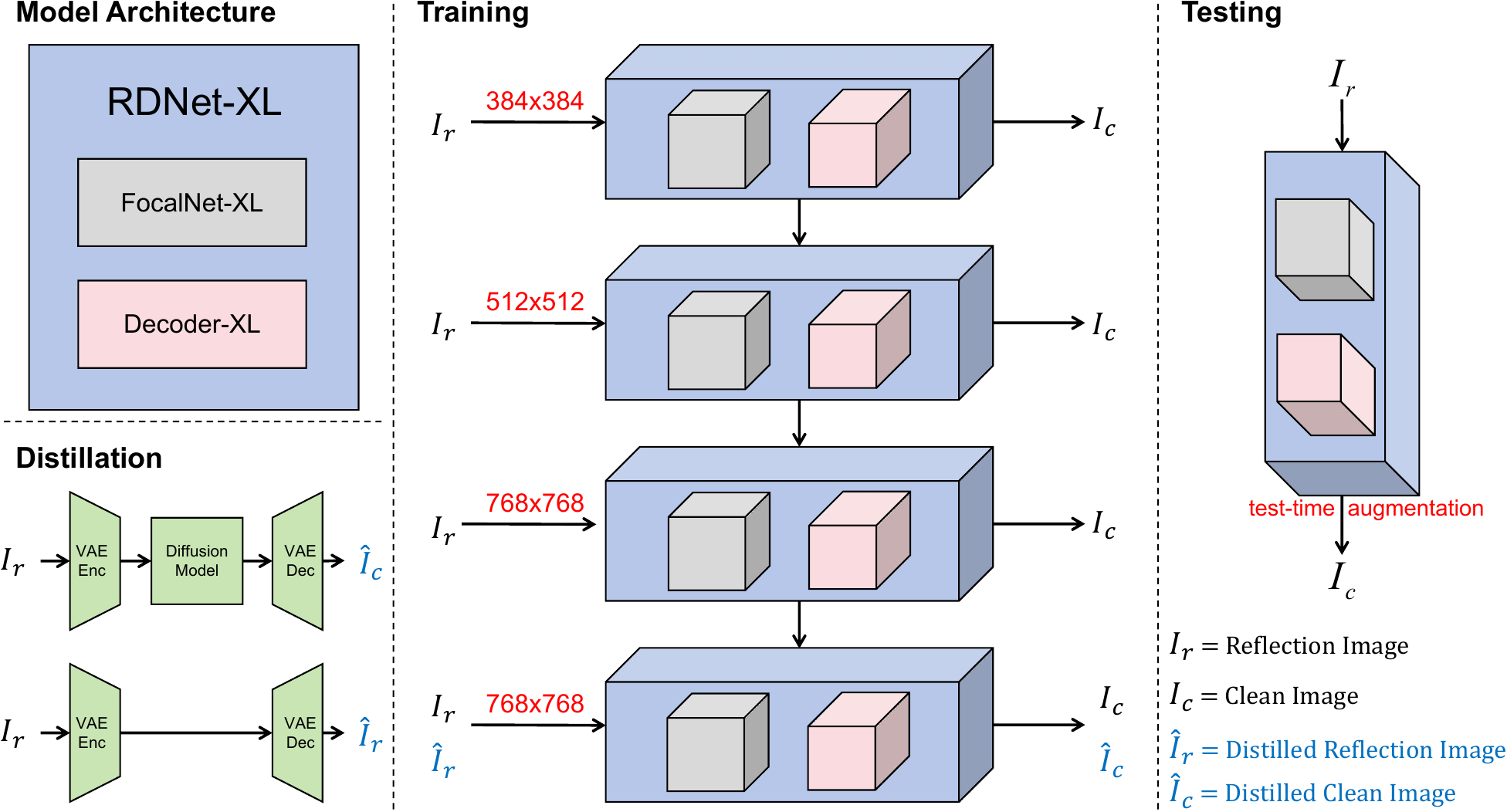}
    \caption{Overview of the proposed RDNet-XL framework with diffusion-based distillation.}
    \label{fig:xreflect_master_framework}
\end{figure}

As shown in Fig.~\ref{fig:xreflect_master_framework}, our final submission is based on RDNet~\cite{rdnet} under the XReflection framework, which provides a strong and stable baseline for reflection removal. To further improve the removal performance under complex reflections in the wild, we replace the original FocalNet-L backbone with the larger FocalNet-XL. This enhancement significantly strengthens multi-scale representation capacity and global contextual modeling, leading to more accurate reflection suppression and detail preservation. Despite the increased model size, the additional computational overhead remains moderate and well justified by the consistent gains observed on the validation and test sets.

We construct a large-scale and diverse training set by combining multiple complementary data sources. The provided OpenRR-5k dataset serves as our primary training data. In addition, we incorporate several publicly available paired real-world reflection removal datasets, including RR4K~\cite{chen2024real} (1,230 image pairs), RRW~\cite{zhu2024revisiting} (a curated subset of 3,000 high-quality pairs), the Perceptual Reflection Removal dataset~\cite{zhang2018single} (109 real pairs), and DRR~\cite{hu2026dereflection} (3,000 carefully selected well-aligned pairs).

To further enhance robustness under extreme and rare reflection cases, we construct an additional hard-case dataset consisting of 1,000 challenging reflection image pairs. Inspired by the strong generative capability of diffusion-based models in recent visual restoration tasks, we employ state-of-the-art diffusion-based reflection removal methods (\eg, WindowSeat~\cite{zakarin2025reflection} and DAI~\cite{hu2026dereflection}) to generate reflection-free pseudo ground truth for reflection images collected from large-scale open-source image datasets. This aims to distill diffusion models' powerful generative priors into our model.
To mitigate reconstruction artifacts and domain gaps introduced by the VAE encoding–decoding process used in diffusion models, we additionally pass each reflection image through the same VAE encoder–decoder pipeline and use the reconstructed image as the network input. This design ensures better domain alignment between the input images and the diffusion-generated supervision.

We adopt a progressive resolution training strategy to stabilize optimization and improve high-resolution performance. Specifically, the model is trained sequentially at image resolutions of 384, 512, and 768 patch size. We train the model for 200 epochs at both 384 and 512 resolutions, followed by 100 epochs at 768 resolution. 
Starting from the 768-resolution checkpoint, we further introduce a knowledge distillation stage using a diffusion-based state-of-the-art reflection removal model as the teacher. During this phase, the synthesized hard-case data are mixed with the original training datasets, and the model is trained for an additional 20 epochs. All experiments are conducted on 8 NVIDIA A100 GPUs using bf16 mixed-precision training, with a per-GPU batch size of 2. We employ the AdamW optimizer with an initial learning rate of $2\times10^{-4}$, which is decayed by a factor of 0.5 every 50 epochs. During inference, we apply test-time augmentation by horizontal flip, vertical flip, and their combination, and average the predictions to further improve robustness and final leaderboard performance.

\subsection{AIIALab -- MS-RDNet: A Multi-Stage Refinement RDNet with Adversarial Perception and Depth-Consistency Scoring}

\begin{figure}[ht]
    \centering
    \includegraphics[width=1.0\linewidth]{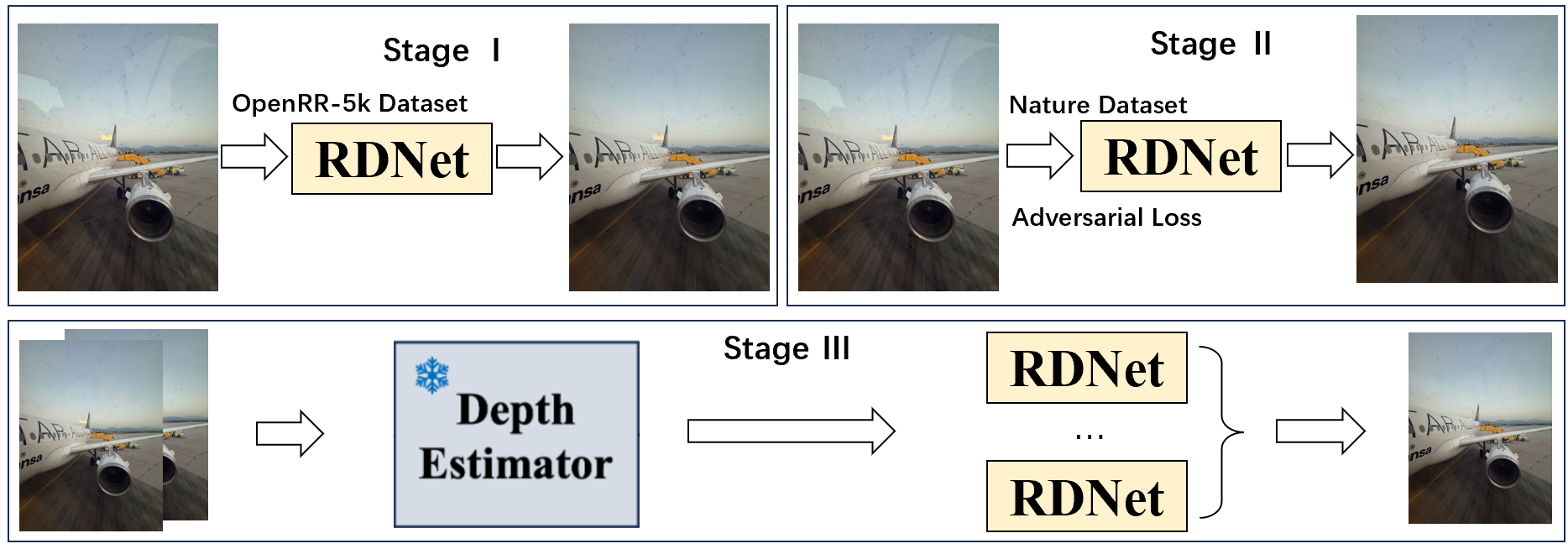}
    \caption{An overview of MS-RDNet for SIRR. Code Link: \href{https://github.com/BrewieRain/MS-RDNet}{https://github.com/BrewieRain/MS-RDNet}}
    \label{fig:framework}
\end{figure}

As shown in Fig.~\ref{fig:framework}, we propose MS-RDNet, a refinement version of RDNet~\cite{zhao2025reversible} with adversarial perception and depth-consistency scoring. 
In the first stage, the RDNet backbone is supervisedly pre-trained on the OpenRR-5k~\cite{cai2025openrr} dataset to establish a robust baseline for transmission and reflection layer separation. To further bridge the gap between synthetic and real-world distributions, the second stage employs the Nature dataset~\cite{li2020single} for perceptual fine-tuning, where an Adversarial Loss is integrated to shift the optimization objective from pixel-level constraints toward human-centric visual fidelity and high-frequency texture recovery. In the third stage, inspired by the depth-consistency scoring strategy from GenSIRR~\cite{li2025rectifying}, we introduce a physical evaluation mechanism that utilizes a depth estimator to score the results generated across different training phases. Base on this, we employ the model merging technology to obtain a more balanced model.

Beyond the competition-provided training dataset, we supplement our training data with 200 additional real pairs from the ``Nature''~\cite{li2020single} dataset.

Implemented using the PyTorch and XReflection frameworks, our model is trained with the AdamW optimizer (initial learning rate = 0.0001, halved every 80,000 iterations) on 2 NVIDIA GeForce RTX 3090 GPUs. We first train on 512×512 random-cropped patches for 100,000 iterations with MSE, gradient, and VGG perceptual losses, then increase the patch size to 784 and add adversarial loss for an additional 60,000 iterations.

\subsection{VIP Lab -- Complementary Mixture-of-Experts and Complementary Cross-Attention for Single Image Reflection Separation in the Wild}

\begin{figure}[ht]
    \centering
    \includegraphics[width=1.0\linewidth]{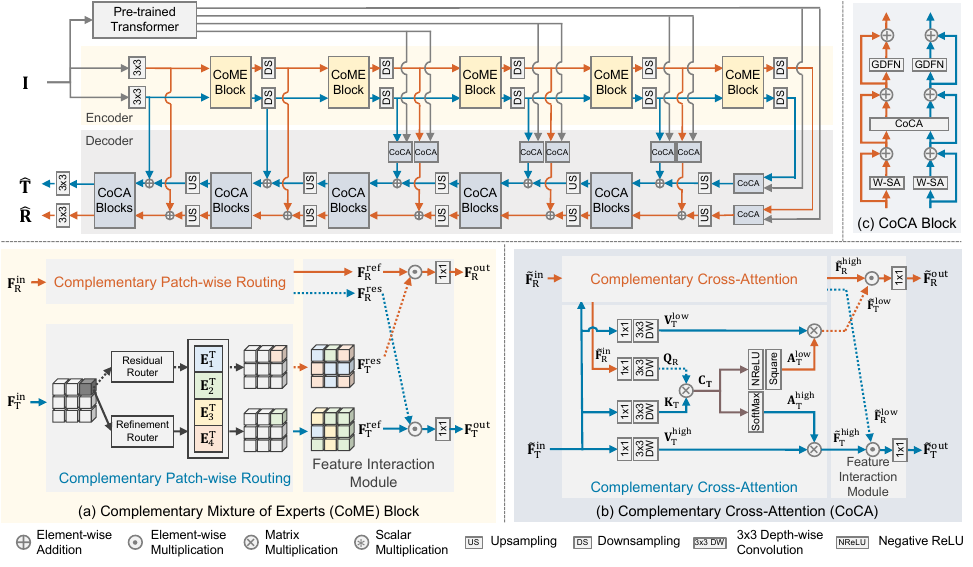}
    \caption{The overall architecture of the proposed network. Code Link: \href{https://github.com/j0nghyukpark/CompSIRS}{github.com/j0nghyukpark/CompSIRS}}
    \label{fig:viplab_arch}
\end{figure}

The overall architecture of the proposed model is shown in Figure~\ref{fig:viplab_arch}, which is based on our prior work~\cite{park2026complementary}, a dual-stream Single Image Reflection Separation framework consisting of Complementary Mixture-of-Experts (CoME) and Complementary Cross-Attention (CoCA).

We adopt a two-stage training strategy. In the first stage, we train our network on $384\times384$ random-cropped patches for 40 epochs using 7,643 synthetic pairs from PASCAL VOC~\cite{everingham2010pascal}, 90 real pairs from~\cite{zhang2018single}, and 200 real pairs from~\cite{li2020single}.
In the second stage, we fine-tune the model on the official competition dataset, OpenRR-5k~\cite{cai2025openrr}. To accommodate its higher resolution compared to the open datasets, we increase the patch size to $768\times768$. We train the model for 20 epochs using 4,950 training pairs, reserving the remaining 50 pairs for validation.

All training is conducted on a single NVIDIA RTX PRO 6000 Blackwell GPU. We employ the Adam optimizer with $\beta_1 = 0.9$, $\beta_2 = 0.999$, a fixed learning rate of $10^{-4}$, and a batch size of 1. We adopt a composite loss function consisting of the reconstruction loss, gradient loss, feature loss, and load balancing loss, following~\cite{park2026complementary}. For testing, we further apply test-time augmentation (TTA) with three inputs: the original image and its horizontal and vertical flips. We perform inference on each of the three inputs and average the results to improve the final output quality.

\subsection{YuFans -- Frequency-Aware Fine-Tuning with Post-Training Optimization for Reflection Removal}

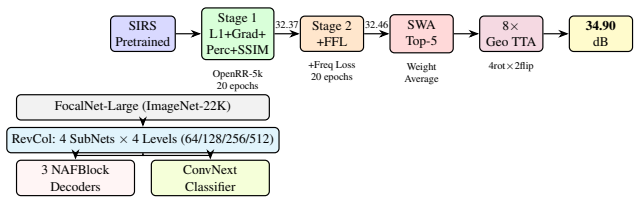
\begin{figure}[ht]
    \centering
    \resizebox{\linewidth}{!}{%
    \begin{tikzpicture}[
      node distance=0.35cm and 0.4cm,
      box/.style={draw, rounded corners=3pt, minimum height=0.8cm, minimum width=1.5cm, font=\small, align=center, thick},
      arrow/.style={-{Stealth[length=2.5mm]}, thick},
      lbl/.style={font=\scriptsize, align=center},
    ]
      \node[box, fill=blue!15] (pretrain) {SIRS\\Pretrained};
      \node[box, fill=green!15, right=0.6cm of pretrain] (s1) {Stage 1\\L1+Grad+\\Perc+SSIM};
      \node[box, fill=orange!20, right=0.6cm of s1] (s2) {Stage 2\\+FFL};
      \node[box, fill=red!15, right=0.6cm of s2] (swa) {SWA\\Top-5};
      \node[box, fill=purple!15, right=0.6cm of swa] (tta) {8$\times$\\Geo TTA};
      \node[box, fill=yellow!25, right=0.6cm of tta] (out) {\textbf{34.90}\\dB};

      \draw[arrow] (pretrain) -- (s1);
      \draw[arrow] (s1) -- node[above, lbl] {32.37} (s2);
      \draw[arrow] (s2) -- node[above, lbl] {32.46} (swa);
      \draw[arrow] (swa) -- (tta);
      \draw[arrow] (tta) -- (out);

      \node[lbl, below=0.1cm of s1] {OpenRR-5k\\20 epochs};
      \node[lbl, below=0.1cm of s2] {+Freq Loss\\20 epochs};
      \node[lbl, below=0.1cm of swa] {Weight\\Average};
      \node[lbl, below=0.1cm of tta] {4rot$\times$2flip};

      \node[lbl, below=0.7cm of pretrain, xshift=1cm] (arch) {};
      \node[box, fill=gray!10, below=1.0cm of pretrain, minimum width=6cm, minimum height=0.6cm] (backbone) {FocalNet-Large (ImageNet-22K)};
      \node[box, fill=cyan!10, below=0.15cm of backbone, minimum width=6cm, minimum height=0.6cm] (revcol) {RevCol: 4 SubNets $\times$ 4 Levels (64/128/256/512)};
      \node[box, fill=pink!15, below=0.15cm of revcol, minimum width=2.8cm, minimum height=0.6cm, xshift=-1.6cm] (dec) {3 NAFBlock\\Decoders};
      \node[box, fill=lime!15, below=0.15cm of revcol, minimum width=2.8cm, minimum height=0.6cm, xshift=1.6cm] (cls) {ConvNext\\Classifier};

      \draw[arrow] (backbone) -- (revcol);
      \draw[arrow] (revcol.south) -- ++(0,-0.05) -| (dec.north);
      \draw[arrow] (revcol.south) -- ++(0,-0.05) -| (cls.north);
    \end{tikzpicture}%
    }
    \caption{Overview of our approach. \textbf{Top:} Two-stage fine-tuning pipeline with SWA and geometric TTA. Numbers on arrows indicate validation PSNR at each stage. \textbf{Bottom:} RDNet architecture with FocalNet-Large backbone, RevCol body, and three separate NAFBlock decoders. Code Link: \href{https://drive.google.com/drive/folders/1wJr9Vm0bwArph8mShwd5GYvLFWG7fxvy}{Google Drive}}
    \label{fig:pipeline}
\end{figure}

We build upon RDNet~\cite{rdnet}, a Reversible Column Network (RevCol) originally pretrained on the SIRS synthetic dataset. The architecture comprises a FocalNet-Large~\cite{yang2022focal} backbone (ImageNet-22K pretrained), 4~SubNets with 4~hierarchical levels (64/128/256/512 channels), 3~separate NAFBlock~\cite{chen2022nafnet} decoders for reflection decomposition, and a frozen ConvNext classifier that provides prompt-based feature modulation. As shown in Fig.~\ref{fig:pipeline}, we adopt a \textbf{two-stage fine-tuning strategy}: Stage~1 adapts the SIRS-pretrained model to the real-world OpenRR-5k distribution using spatial losses ($\mathcal{L}_1 + 0.1\mathcal{L}_{\text{grad}} + 0.01\mathcal{L}_{\text{VGG19}} + 0.1\mathcal{L}_{\text{SSIM}}$), reaching 32.37~dB validation PSNR after 20~epochs. Stage~2 introduces \textbf{Focal Frequency Loss (FFL)}~\cite{jiang2021focal} with weight 0.01, which adaptively emphasizes frequency bands where the model underperforms --- this is particularly effective because residual reflections tend to concentrate in specific frequency ranges that spatial losses fail to adequately penalize.

We train exclusively on the OpenRR-5k dataset (5,000 pairs) with \textbf{no external data}. Our augmentation consists of random horizontal/vertical flips and 90\textdegree/180\textdegree/270\textdegree~rotations. Notably, we found that adding synthetic SIRS data \emph{degrades} performance (val PSNR drops from 32.77 to 31.94) due to domain gap, and MixUp augmentation also hurts by destroying the physical blended-transmission relationship. We use AdamW (weight decay $10^{-4}$) with differential learning rates: backbone 5e-6, decoder heads 2e-5, classifier 2e-4. Training uses bfloat16 mixed precision with gradient accumulation (effective batch size 8) and MultiStepLR schedule ($\gamma$=0.5 at 30\%/60\%/90\% of training).

Our post-training pipeline contributes substantially to final performance. We apply \textbf{Stochastic Weight Averaging (SWA)}~\cite{izmailov2018swa}, selecting the top-5 epoch checkpoints by validation PSNR and averaging their weights, which produces smoother and more robust predictions (+0.05~dB over the best single checkpoint on Codabench). At inference, we employ \textbf{8$\times$ geometric TTA} (4~rotations $\times$ 2~flips), consistently boosting PSNR by $\sim$1.8~dB. We also explored Model Soup (weight-space averaging across different random seeds), achieving 34.88~dB, but SWA within a single well-trained run proved slightly more effective (34.90~dB). Critically, we discovered that multiscale TTA is \emph{catastrophically harmful} ($-$2.48~dB), as rescaling introduces interpolation blur that degrades reflection removal quality. Our final submission uses the same SWA checkpoint for both validation and test sets.

\subsection{KLETech-CEVI -- XReflection-based Deep Network for Single Image Reflection Removal}

\begin{figure}[ht]
\centering
\includegraphics[width=\linewidth]{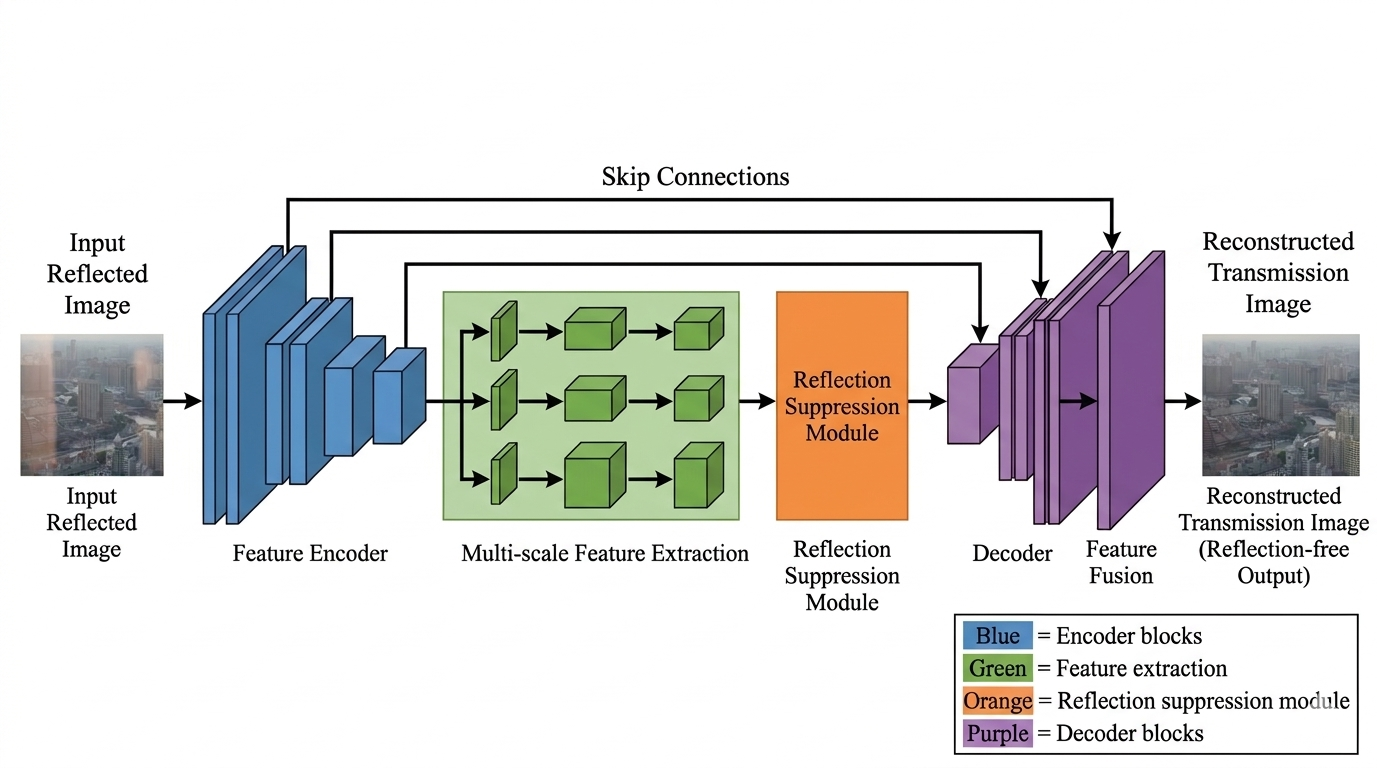}
\caption{Overview of the proposed reflection removal framework based on the XReflection architecture. The network learns to recover the clean transmission layer from a reflection-contaminated input image through hierarchical feature extraction and residual reflection estimation. Code Link: \href{https://github.com/CeviKle/NTIRE2026-KLETech-CEVI-SIRR.git}{https://github.com/CeviKle/NTIRE2026-KLETech-CEVI-SIRR.git}}
\end{figure}

Our method adopts a hierarchical encoder--decoder architecture inspired by the XReflection framework. The architecture is designed to effectively capture both global scene context and fine-scale reflection patterns.

The encoder progressively extracts hierarchical features from the input image using stacked convolutional layers. Each stage performs spatial downsampling while increasing the number of feature channels. This process enables the network to encode high-level contextual information and reflection characteristics at different scales.

The decoder reconstructs the clean transmission layer through a sequence of upsampling and convolution operations. Skip connections are introduced between corresponding encoder and decoder layers to preserve spatial information and improve gradient propagation during training.

Instead of directly predicting the clean transmission layer, the network is trained to estimate the reflection component. The final transmission image is obtained by subtracting the predicted reflection from the input image:

\begin{equation}
\hat{T} = I - F_{\theta}(I)
\end{equation}

This residual formulation simplifies the learning objective and allows the network to focus specifically on reflection artifacts.

Reflections in real-world images may appear at different spatial scales and intensities. To effectively handle such variations, the network aggregates features across multiple scales. This hierarchical representation enables the model to capture both large reflection structures and subtle reflection patterns.

The final training objective is defined as a weighted combination of the above losses:

\begin{equation}
L_{total} = \lambda_1 L_{pixel} + \lambda_2 L_{perc} + \lambda_3 L_{ssim}
\end{equation}

where $\lambda_1$, $\lambda_2$, and $\lambda_3$ balance the contribution of each loss component.

\subsection{PSU -- DUSKAN: Dual Spectral Kolmogorov-Arnold Network}

\begin{figure}[ht]
    \centering
    \includegraphics[width=1.0\linewidth]{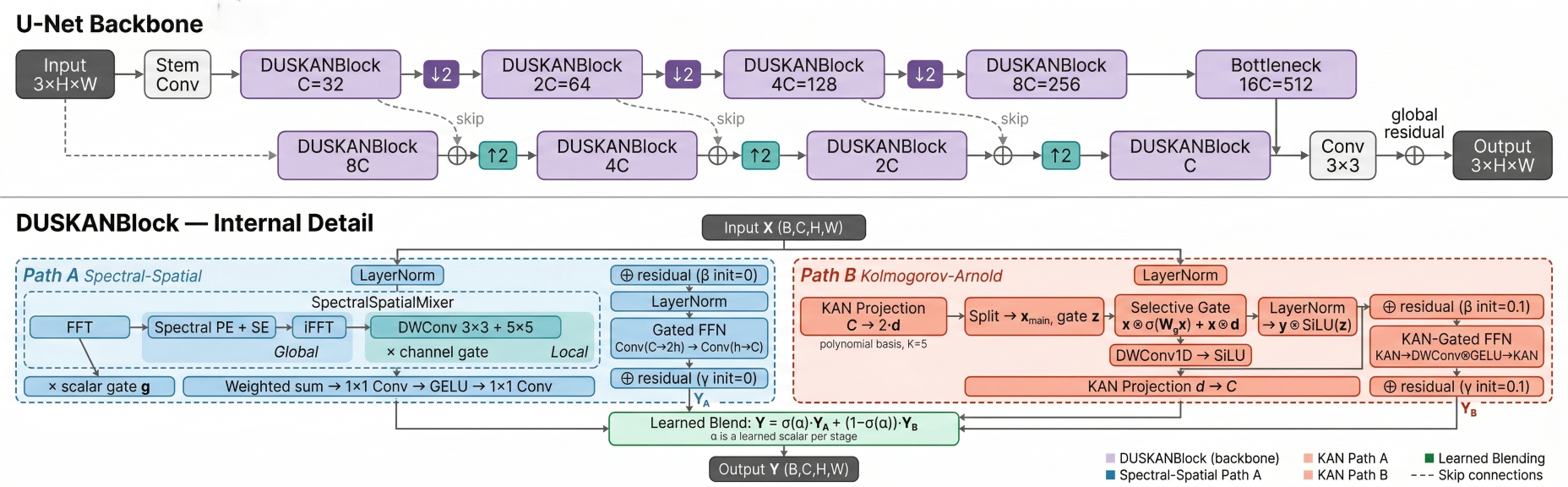}
    \caption{\textbf{DUSKAN architecture.} Symmetric 4-level U-Net with DUSKANBlock stages. Path A (blue) extracts global features via FFT magnitude modulation, enriched with spectral positional encoding and SE reweighting. Path B (red) uses Kolmogorov-Arnold polynomial-basis activations with a parallel-additive selective gate. A learned per-stage logit $\alpha$ blends both outputs. Code Link: \href{https://drive.google.com/drive/folders/1ud-V4YNdWMaoKGYT3lBJ40EiHaLBc1gn?usp=sharing}{Google Drive}}
    \label{fig:duskan_arch}
\end{figure}

As shown in Fig.~\ref{fig:duskan_arch}, we propose DUSKAN (Dual Spectral Kolmogorov-Arnold Network), a dual-path architecture designed to balance global degradation removal and fine local texture preservation. The model features a symmetric 4-level U-Net backbone where each stage acts as a DUSKANBlock. The spectral-spatial path (Path A) handles global frequency shifts caused by artificial light and haze via 2D FFT magnitude modulation, spatial positional encodings, and multi-scale depthwise convolutions. Parallel to this, the adaptive path (Path B) utilizes Kolmogorov-Arnold Networks (KAN) to learn data-driven, polynomial-basis activation functions per edge, effectively managing spatially varying degradations. The outputs of both paths are dynamically blended using a learned per-stage gating weight, optimizing the representation at each scale.

We strictly utilized only the official training and validation datasets provided by the NTIRE 2026 Reflection Removal in the Wild Challenge. No external datasets, pre-trained weights, or extra synthesized pairs were used. During training, we extracted random patches of size $512 \times 512$ and applied standard geometric data augmentations, including random rotations ($0^\circ$, $90^\circ$, $180^\circ$, $270^\circ$) and horizontal/vertical flips to prevent overfitting and improve generalization.

The network was implemented using PyTorch (v2.2) and trained from scratch on a single NVIDIA A100 (80GB) GPU. We optimized the model using the AdamW optimizer (initial learning rate of $2 \times 10^{-4}$, decaying to $1 \times 10^{-6}$ via a Cosine Annealing scheduler) for 500 epochs with a batch size of 2, utilizing Automatic Mixed Precision (AMP) for efficiency. The training objective combines L1, VGG perceptual, and focal frequency losses. For inference, DUSKAN processes the full-resolution images directly in a single forward pass without the need for overlapping patches or test-time augmentation.

\subsection{SiGMoid -- Two-head Restormer with checkpoint soup and horizontal flip TTA}

\begin{figure}[t]
    \centering
    \includegraphics[width=0.58\linewidth]{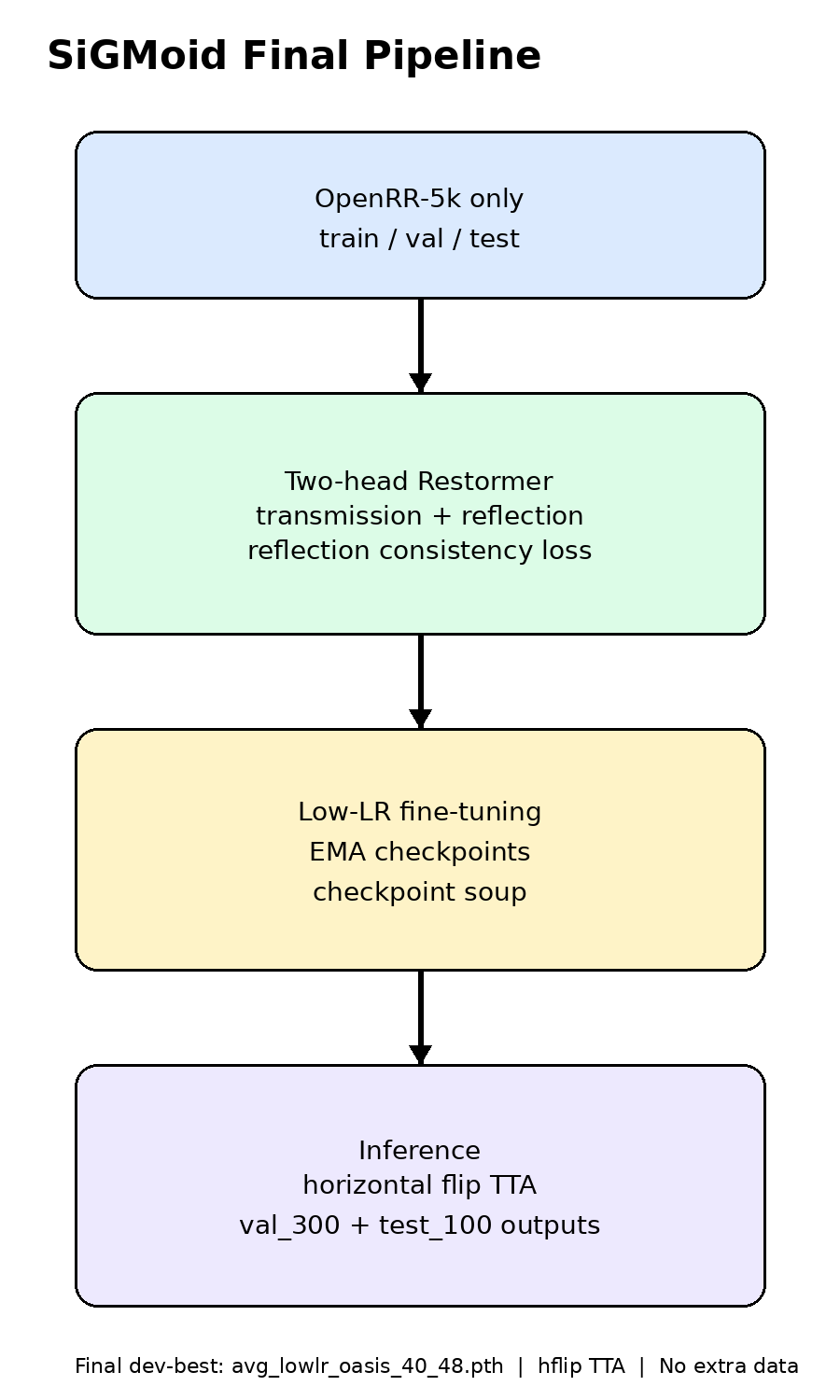}
    \caption{Overview of the final SiGMoid pipeline used for the NTIRE 2026 SIRR final submission. Code Link: \href{https://drive.google.com/file/d/1380Rnw3Qn39FrypZJuEd7ebzMnrkt3Ol/view?usp=sharing}{Google Drive}}
    \label{fig:sigmoid_pipeline}
\end{figure}

As shown in Fig.~\ref{fig:sigmoid_pipeline}, our final model uses a 4-stage Restormer with $dim=48$, blocks $[4,6,6,8]$, heads $[1,2,4,8]$, and 6 output channels. The first three channels predict the transmission layer and the last three channels predict the reflection component. This two-head design is our main architectural modification over the plain baseline: the network is asked to explicitly explain the input image as a sum of clean content and reflection, instead of producing only one restored output.

We supervise the transmission head with an $L_1$ reconstruction loss. For the reflection head, we use a reflection loss with $\lambda_r=0.2$ against a proxy reflection target derived from the input and ground truth. We also use a transmission-reflection consistency term with $\lambda_c=0.05$ so that the two outputs reconstruct the observed blended image in a physically meaningful way. In practice, this decomposition loss is more stable than adding heavy perceptual or adversarial objectives, and it kept training reliable under limited experiment time.

We use only the challenge-provided OpenRR-5k train split and no extra data. Training uses random 512$\times$512 patches, batch size 1, AdamW, cosine scheduling, mixed precision, and EMA with decay 0.999. We apply lightweight augmentation only: horizontal flip, small rotation and translation, color jitter, and gamma perturbation. We intentionally kept augmentation moderate because stronger geometric or synthetic-domain perturbations tended to move the samples away from the reflection patterns seen in the official validation set.

The main training run uses learning rate $2\times10^{-4}$ for 200 epochs. After convergence, we continue from the trained weights with a low learning rate of $1\times10^{-5}$ to refine the model. This continuation stage is important in our pipeline: it improves stability and provides several nearby late checkpoints that can later be merged. We use the official validation split only for model selection.

Our final submitted model is not a single late epoch but an equal-weight checkpoint soup of three models: the best checkpoint from the low-learning-rate fine-tuning run and two stable continuation snapshots (epochs 40 and 48 of the continuation run). This simple averaging was one of the most reliable practical gains in the final stage of the competition, while additional long continuation without averaging showed little further improvement.

The same averaged checkpoint is used for both validation and test submissions, following the challenge requirement. During inference we do not resize the input image. Instead, we pad the image to a multiple of 8, restore it, and crop it back to the original size. We then apply horizontal-flip test-time augmentation and average the original and flipped predictions. We tested heavier TTA variants such as rotation-based D4 TTA and multi-scale inference, but they did not improve the validation score, so the final system keeps only horizontal flip.

The final dev-best checkpoint is \texttt{avg\_lowlr\_oasis\_40\_48.pth}. It uses no extra data and runs on a single RTX 6000 Ada 48GB GPU. This setting achieved our best CodeBench validation result: PSNR 34.246, LPIPS 0.0296, and SSIM 0.9747. The same checkpoint and inference setting were used to prepare both the final validation outputs and the final test outputs.

\subsection{NTR -- TimeDiffiT}

We adopt a two-stage training strategy: (1)~self-supervised pretraining via Masked Diffusion Autoencoding (MDAE)~\cite{tu2025score,he2022masked}, followed by (2)~supervised fine-tuning on the challenge dataset using the Diffusion-to-Score SFT (D2S-SFT) loss~\cite{tu2025score}.

\textbf{Stage~1: MDAE Pretraining.}
The network is pretrained on large-scale image corpora via MDAE~\cite{tu2025score,he2022masked}, which applies dual corruption: (1)~spatial masking of $p_{\text{mask}} \sim \mathcal{U}(1\%, 75\%)$ of non-overlapping $16{\times}16$ blocks, and (2)~VE-SDE noise injection~\cite{song2021scorebased}: $\tilde{X}_t = X_0 + \sigma_t Z$, $Z \sim \mathcal{N}(\mathbf{0}, \mathbf{I})$. The MDAE loss combines an MAE reconstruction term~\cite{he2022masked} with a Corruption2Self (C2S) self-supervised diffusion matching term~\cite{tu2025score}, enabling label-free pretraining.

\textbf{Stage~2: D2S-SFT Fine-Tuning.}
We fine-tune the full pretrained encoder-decoder on 4,750 paired training images (blended / transmission layer) from the challenge dataset. We set the diffusion time conditioning to $t = 0$ at both training and inference, as blended reflection images contain no additive Gaussian noise. Training uses the D2S-SFT loss~\cite{tu2025score} with the AdamW optimizer~\cite{loshchilov2019adamw} (lr $= 2 \times 10^{-4}$, cosine decay), batch size 8 per GPU, and random $512 \times 512$ crops. We train for 300 epochs, then refine for 90 additional epochs incorporating the full training set.

\begin{figure}[ht]
\centering
\includegraphics[width=\linewidth]{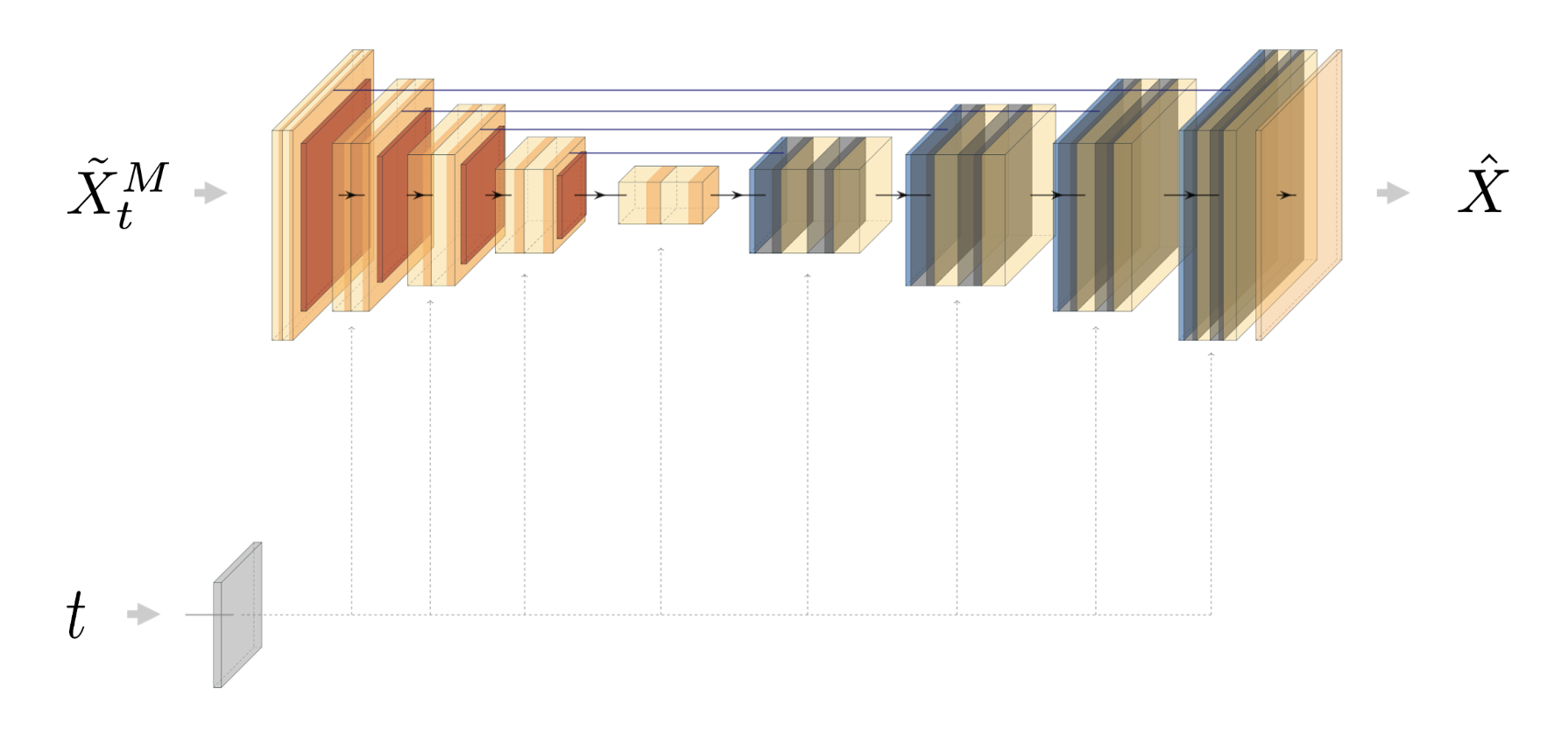}
\caption{TimeDiffiT architecture. The doubly-corrupted input $\tilde{X}_t^M$ and noise level $\sigma_t$ enter the time-conditioned U-Net (encoder: orange; decoder: blue), producing the restored output $\hat{X}$. During SFT, no masking is applied and $\sigma_t = 0$.}
\label{fig:arch}
\end{figure}

We use TimeDiffiT~\cite{tu2025score}, a time-conditional U-Net with 142.5M parameters (Fig.~\ref{fig:arch}). The architecture has 4 spatial scales with channel dimensions $[128, 256, 512, 1024]$. Time conditioning is realized via a scalar $\sigma_t$ encoded through a sinusoidal embedding and 2-layer MLP into a 512-d vector, which modulates each ResNet block via Adaptive Group Normalization (AdaGN). Shifted-window TimeAttention blocks operate at all 4 encoder/decoder levels plus the bottleneck (9 blocks total, 4 heads, 32 dimensions per head).

For reflection removal, input and output are both 3-channel RGB. The full encoder-decoder is fine-tuned end-to-end (not encoder-only as in the standard MDAE representation learning setup).

\subsection{refineX -- Progressive Restormer U-Net with Multi-Loss Training for Single Image Reflection Removal}

\begin{figure}[ht]
    \centering
    \includegraphics[width=1.0\linewidth]{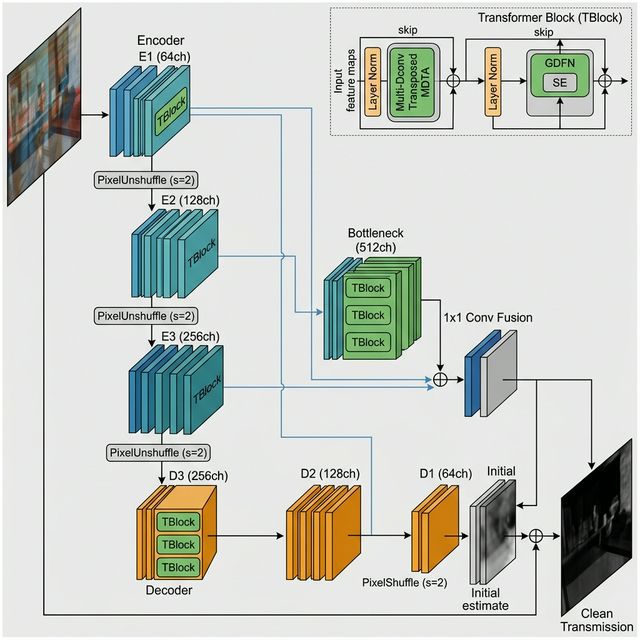}
    \caption{Overview of our proposed SIRR-Net for single image reflection removal.
             The U-Net encoder-decoder processes the blended input through four resolution
             levels, with each stage comprising stacked Transformer Blocks (MDTA + GDFN + SE).
             Skip connections fuse encoder and decoder features.
             A global residual connection adds the input to the output. Code Link: \href{https://github.com/Shreenikethjoshi/NTIRE2026-Reflection-removal-in-the-wild}{https://github.com/Shreenikethjoshi/NTIRE2026-Reflection-removal-in-the-wild}}
    \label{fig:architecture}
\end{figure}

As illustrated in Fig.~\ref{fig:architecture}, we propose \textbf{SIRR-Net}, a Restormer-style
hierarchical U-Net designed for single image reflection removal.
The network follows a symmetric encoder-decoder structure operating at four resolution scales.

\noindent\textbf{Encoder.}
The input blended image is first projected to $D{=}64$ feature channels via a $3{\times}3$ convolution.
The encoder comprises three stages:
\textbf{E1} ($64$\,ch, $4$ blocks),
\textbf{E2} ($128$\,ch, $6$ blocks), and
\textbf{E3} ($256$\,ch, $6$ blocks),
each connected by a pixel-unshuffle-based downsampling layer that halves spatial
resolution without information loss.
A bottleneck stage (\textbf{Bot}, $512$\,ch, $8$ blocks) processes the deepest representation.

\noindent\textbf{Decoder.}
Symmetric decoder stages \textbf{D3}--\textbf{D1} upsample features via pixel-shuffle.
Skip connections from the encoder are concatenated with the upsampled features
and fused through a $1{\times}1$ convolution before each decoder stage.
A final $3{\times}3$ convolution maps back to 3 channels, and a
\textbf{global residual connection} adds the input image to produce the clean output:
\begin{equation}
  \hat{T} = \text{clamp}\!\left(\text{Dec}(\text{Enc}(I)) + I,\; 0, 1\right).
\end{equation}

\noindent\textbf{Transformer Block (TBlock).}
Each processing stage is composed of stacked TBlocks, each containing:
\begin{itemize}
  \item \textbf{MDTA} -- Multi-Dconv Head Transposed Attention~\cite{restormer}:
        channel-wise attention with complexity $\mathcal{O}(C^2)$ instead of
        $\mathcal{O}((HW)^2)$, enabling high-resolution training.
        Depthwise convolutions capture local context before attention.
  \item \textbf{GDFN} -- Gated Depthwise Feed-Forward Network~\cite{restormer}:
        uses a SimpleGate (element-wise product of two parallel branches)
        for non-linear feature modulation with expand ratio $2.66\times$.
  \item \textbf{SE} -- Squeeze-and-Excite block applied after every stage
        to recalibrate channel-wise importance using both average and
        max-pool global context.
\end{itemize}
LayerNorm (reimplemented in 2D) is used for stable training with AMP.
Gradient checkpointing is applied to the bottleneck during training to reduce
GPU memory consumption.

We employ a three-stage progressive training schedule to stabilize
optimization from coarse to fine detail:

\begin{enumerate}
  \item \textbf{Stage 1 -- Coarse (128\,px):}
        Initial convergence on small patches with standard loss weights.
        Pre-trained checkpoint $\mathbf{w}_1$ is provided externally.
  \item \textbf{Stage 2 -- Mid (256\,px, 60 epochs):}
        Training at medium resolution with
        $\text{lr}{=}5{\times}10^{-5}$, batch size $6$ per GPU.
        Initialized from Stage~1 weights $\mathbf{w}_1$.
  \item \textbf{Stage 3 -- Perceptual (384\,px, 30 epochs):}
        Fine-tuning at higher resolution with increased perceptual and
        edge loss weights to improve visual quality for subjective evaluation.
        lr${}=10^{-5}$, batch size $3$ per GPU.
\end{enumerate}

All stages use \textbf{DistributedDataParallel (DDP)} via \texttt{torchrun}
on 3$\times$ NVIDIA RTX 6000 Ada (48\,GB each).
The learning rate is scaled as $\text{lr}_{\text{eff}} = \text{lr} \cdot \sqrt{B_{\text{eff}}/8}$
and annealed with \textit{CosineAnnealingWarmRestarts}.
Mixed-precision (AMP fp16) and gradient clipping ($\|\nabla\|_2 \leq 0.01$) are used throughout.

\subsection{ACVLAB -- RDNet with Frozen DINOv2 Semantic Prior for Reflection Removal}

\begin{figure}[ht]
    \centering
    \includegraphics[width=1.0\linewidth]{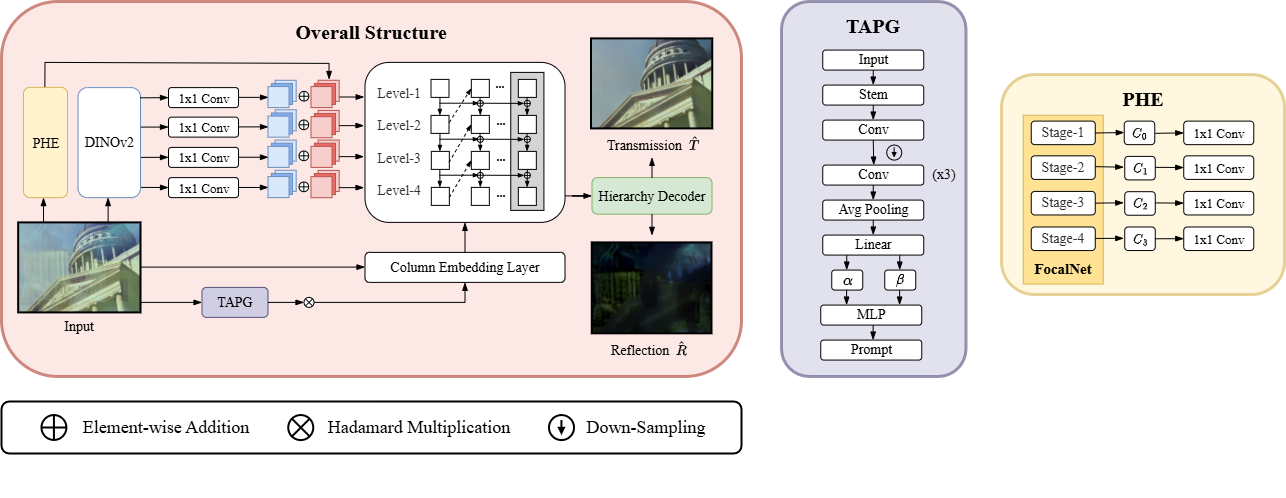}
    \caption{An overview of RDNet+ for SIRR. Code Link: \href{https://github.com/wuw2135/RDNet-with-Frozen-DINOv2-Semantic-Prior-for-Reflection-Removal}{https://github.com/wuw2135/RDNet-with-Frozen-DINOv2-Semantic-Prior-for-Reflection-Removal}}
    \label{fig:x_reflection_pipeline}
\end{figure}

Our method builds upon RDNet~\cite{rdnet}, which consists of a multi-column reversible encoder (MCRE), a FocalNet-L-based pretrained hierarchy extractor (PHE), and a transmission-rate-aware prompt generator (TAPG). We observe that under strong specular reflections the original model tends to produce blurry reconstructions, as FocalNet features alone lack sufficient object-level semantics to infer plausible appearance after reflection removal. To address this, we introduce a frozen DINOv2~\cite{dinov2} ViT-B/14 branch as an auxiliary semantic prior: the input is resized to $518\!\times\!518$ and fed into the frozen encoder; patch tokens are reshaped into a 2-D feature map and projected through four $1\!\times\!1$ convolutional adapters to match each MCRE level (64, 128, 256, 512 channels), then bilinearly interpolated and fused via element-wise addition. With DINOv2~\cite{dinov2} entirely frozen, training cost is limited to the four lightweight adapters while significantly reducing blurring artifacts.

We initialize the main components from RDNet pretrained weights and attach randomly initialized DINOv2~\cite{dinov2} adapters, then fine-tune on OpenRR-1k and OpenRR-5k using a single A100 GPU. We adopt AdamW with a \texttt{ReduceLROnPlateau} scheduler that monitors validation PSNR and decays the learning rate by a factor of 0.85 upon plateau. The loss follows the original RDNet formulation combining MSE content loss, gradient-domain $\ell_1$ loss, and VGG-19 perceptual loss. The two-phase strategy---loading the full RDNet~\cite{rdnet} checkpoint first, then fine-tuning with the DINOv2~\cite{dinov2} branch---preserves the encoder's learned decoupling capability while the added semantic features gradually steer the decoder toward sharper reconstructions.

%% file: Tables/table_3.tex
\begin{sidewaystable*}[htbp]
\centering
\renewcommand{\arraystretch}{1.5} 
\begin{tabular}{clccccc}
\hline
\hline
Team Name & Method & Training Data & Traintime [h] & Runtime [s] & Params [M] & FLOPs [G] \\[1ex]
\hline  \noalign{\smallskip}
RRay            & RDNet + NAFNet  & \shortstack[l]{OpenRR-5k$_{train}$, Real-Nature, \\ RRW, Synthetic Dataset}  & 48 (A100)  & 1.2131 (A10)  & 322.82  & 1934.50 \\[3ex]
Xreflect Master & UNet & \shortstack[l]{OpenRR-5k$_{train}$, RR4K, \\ RRW, DRR, \\ perceptual-reflection-removal} & 180 (8*A100) & 5.6 (A100) & 545.52 & 5175.93  \\[3ex]
AIIALab         & RDNet & OpenRR-5k$_{train}$, Nature & 352 (3090) & 1.6 (3090) & 314 & 7340   \\[3ex]
VIP Lab         & Park and Sim~\cite{park2026complementary} & \shortstack[l]{OpenRR-5k$_{train}$, RRW, \\ Real-Nature, \\ Synthetic Dataset} & 71 (RTX 6000) & 0.988 (RTX 6000) & 404 & 2269   \\[3ex]
YuFans          & \shortstack[l]{RDNet \\ RevCol + FocalNet-Large \\ NAFBlock decoders} & OpenRR-5k$_{train}$ & 12 (H800) & 5.0 (H800) & 522.34 & 671.27   \\[3ex]
KLETech-CEVI    & Encoder--Decoder CNN & OpenRR-5k$_{train}$ & 27 (RTX 6000) & 0.9 (RTX 6000) & 27.4 & —   \\[3ex]
PSU             & \shortstack[l]{UNet + KAN \\ Spectral-Spatial Processing}  & OpenRR-5k$_{train}$ & 3.46 (A100) & 0.74 (A100) & 45.599 & 493.36   \\[3ex]
SiGMoid         & 4-stage two-head Restormer & OpenRR-5k$_{train}$ & 180 (RTX 6000) & 2.36 (RTX 6000) & 26.13 & 3389.21   \\[3ex]
NTR             & \shortstack[l]{TimeDiffiT \\ MDAE + D2S-SFT} & OpenRR-5k$_{train}$ & — & 6.5 (H200) & 142.5 & —   \\[3ex]
refineX         & Restormer-based U-Net & OpenRR-5k$_{train}$ & 18 (3*RTX 6000) & 1.8 (RTX 6000) & 8.2 & 120   \\[3ex]
ACVLAB          & \shortstack[l]{RDNet + DINOv2} & OpenRR-5k$_{train}$ & 35 (A100) & 528.391 (A100) & 267.166 & 5749.151  \\
\hline
\hline
\end{tabular}
\caption{Summary of technical specifications and computational complexity for the participating methods. "Traintime" denotes the total training duration (with hardware in parentheses), and "Runtime" refers to the average inference time per image at approximately 1k ($1024 \times 768$) resolution. Model parameters (Params) and FLOPs are also reported to evaluate computational efficiency. Note that all technical details and performance metrics reported in this table are provided by the respective participating teams. The symbol '—' indicates that the corresponding information was not provided by the participants.}
\label{tab:method_comparison}
\end{sidewaystable*}

%% file: Sections/5_AIGC.tex
\section{Comparison with Generative Methods}
\label{sec:aigc_methods}

\begin{figure*}[ht]
    \centering
    \includegraphics[width=1.0\linewidth]{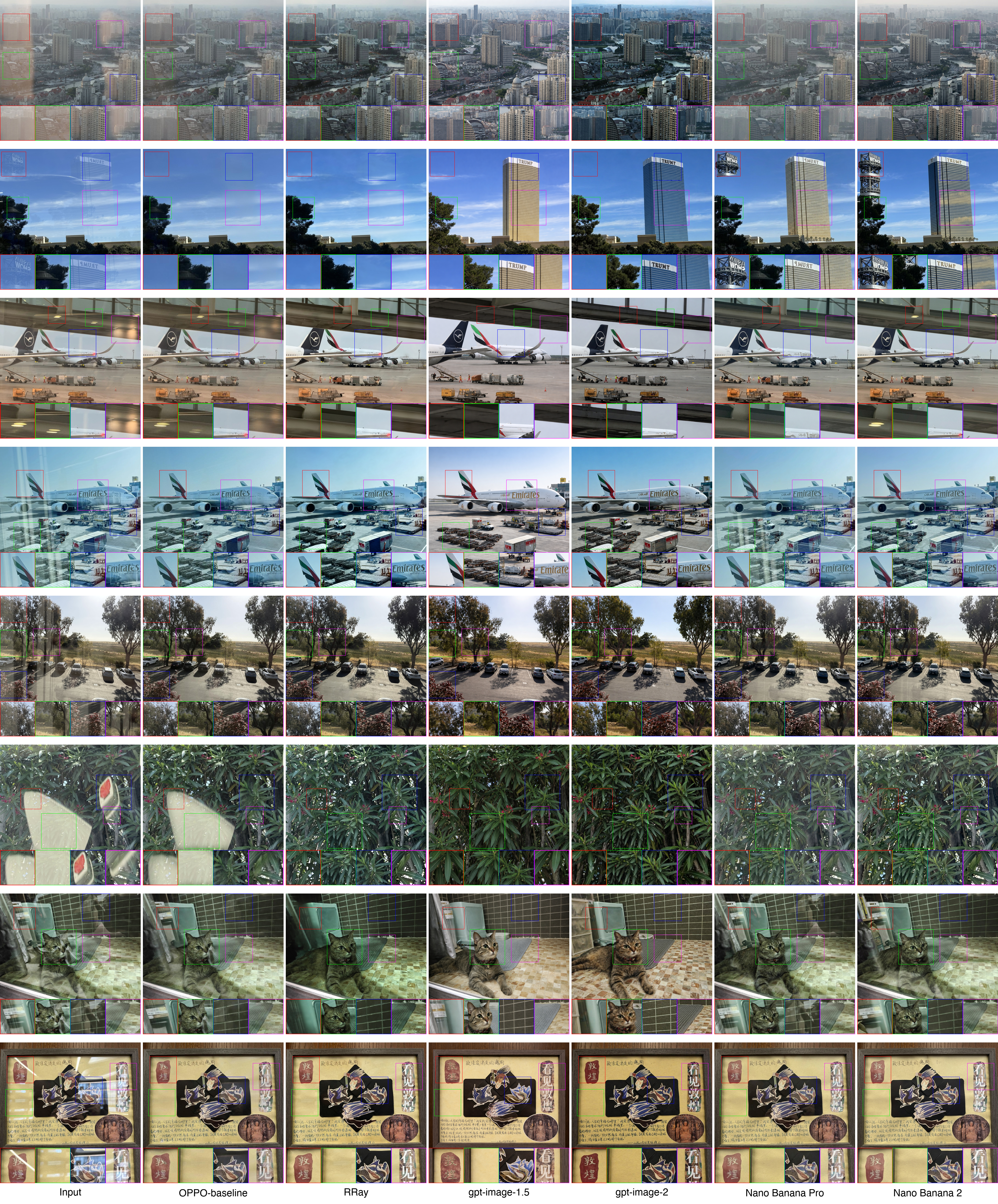}
    \caption{Visual comparison of reflection removal results. From left to right: input images, and results generated by OPPO-baseline, RRay, gpt-image-1.5, gpt-image-2, Nano Banana Pro, and Nano Banana 2. The generative methods (gpt-image series and Nano Banana series) demonstrate superior performance in restoring background details and suppressing artifacts. However, as seen in the second row, generative models may occasionally misidentify intense reflections as background structures, leading to unintended artifact amplification.}
    \label{fig:visual_aigc}
\end{figure*}

In addition to the discriminative approaches proposed by challenge participants, we further explore the potential of generative models for SIRR. Specifically, we evaluate several generative baselines, including \textit{gpt-image} series and \textit{Nano Banana} series. 

As illustrated in Figure~\ref{fig:visual_aigc}, while competition-based methods such as RRay and the OPPO-baseline achieve competitive quantitative scores, they occasionally struggle with complex, non-linear reflections that overlap with high-frequency background textures. In contrast, generative methods leverage strong prior knowledge to "hallucinate" and reconstruct the underlying transmission layer. 

Our observations are as follows:
\begin{itemize}
    \item \textbf{Detail Restoration:} The Nano Banana series and gpt-image-2 demonstrate a superior ability to restore sharp details in the background, effectively suppressing residual ghosting artifacts that often persist in standard CNN or Transformer-based pipelines.
    \item \textbf{Semantic Consistency:} Generative models maintain better semantic integrity in heavily occluded regions, although they may introduce slight stochastic variations that do not always align perfectly with the ground truth pixel-wise.
    \item \textbf{Generalization:} The zero-shot capabilities of models like gpt-image-1.5 suggest that large-scale pre-training on diverse image-text pairs provides a robust prior for decoupling blended layers in real-world scenarios.
    \item \textbf{Pixel Misalignment:} Despite its strong prior, models like gpt-image-1.5 struggle with strict pixel-wise alignment. They often hallucinate textures that, while visually plausible, deviate from the ground truth spatial structures, leading to lower scores in traditional metrics like PSNR.
    \item \textbf{Misidentification and Artifact Amplification:} Generative models can occasionally misinterpret complex reflection patterns as background objects. In certain scenarios, this leads to the unintended amplification of reflections or the introduction of "ghosting" artifacts that were not present in the original input.
\end{itemize}

%% file: Sections/6_Conclusion.tex
\section{Conclusion}
\label{sec:conclusion}

In this paper, we presented the NTIRE 2026 Challenge on Single Image Reflection Removal in the Wild, aiming to advance research on reflection removal in realistic scenarios. We introduced OpenRR-5k, a large-scale dataset consisting of 5,000 training pairs, 300 validation pairs, and 100 test images captured from diverse real-world environments. The challenge attracted strong participation from the research and industry communities and demonstrated the effectiveness of recent deep learning approaches for reflection removal. The results also reveal remaining challenges, particularly in achieving consistent perceptual quality across diverse scenes. We hope that the proposed dataset, benchmark, and analysis will encourage further research and facilitate the development of more robust reflection removal methods for practical applications.

%% file: Sections/Appendix-A.tex
\clearpage  
\begin{appendices}
\section{Teams and Affiliations}
\label{sec:appendix}



\textit{\textbf{Organizers}}:  
\begin{itemize}[leftmargin=*, noitemsep, label={}]
    \item Jie Cai$^1$ (\href{mailto:caijie0620@gmail.com}{caijie0620@gmail.com})  
    \item Kangning Yang$^1$ (\href{mailto:ky189@scarletmail.rutgers.edu}{ky189@scarletmail.rutgers.edu}) 
    \item Zhiyuan Li$^2$ (\href{mailto:zhiyuanli1992@gmail.com}{zhiyuanli1992@gmail.com})
    \item Florin-Alexandru Vasluianu$^{3}$ \newline (\href{mailto:florin-alexandru.vasluianu@uni-wuerzburg.de}{florin-alexandru.vasluianu@uni-wuerzburg.de})
    \item Radu Timofte$^{3}$ (\href{mailto:radu.timofte@uni-wuerzburg.de}{radu.timofte@uni-wuerzburg.de})
    \item Jinlong Li$^1$ (\href{mailto:jinlong.li1@oppo.com}{jinlong.li1@oppo.com})
    \item Jinglin Shen$^1$ (\href{mailto:jinglin.shen@oppo.com}{jinglin.shen@oppo.com})
    \item Zibo Meng$^1$ (\href{mailto:zibo.meng@oppo.com}{zibo.meng@oppo.com})
\end{itemize}

\par\vspace{1pt}

\textit{\textbf{Affiliations}}: 
\begin{itemize}[leftmargin=*, noitemsep, label={}]
    \item $^1$ OPPO, US
    \item $^2$ Meta, US
    \item $^3$ Computer Vision Lab, University of Würzburg, Germany
\end{itemize}

\par\vspace{12pt}

\textbf{Team:} RRay

\textit{\textbf{Members}}:  
\begin{itemize}[leftmargin=*, noitemsep, label={}]
    \item Junyan Cao$^1$ (\href{mailto:junyan_cao@intsig.net}{junyan\_cao@intsig.net})  
    \item Lu Zhao$^1$ (\href{mailto:lu_zhao@intsig.net}{lu\_zhao@intsig.net}) 
    \item Pengwei Liu$^1$ (\href{mailto:pengwei_liu@intsig.net}{pengwei\_liu@intsig.net})
    \item Yuyi Zhang$^1$ (\href{mailto:yuyi_zhang@intsig.net}{yuyi\_zhang@intsig.net})
    \item Fengjun Guo$^1$ (\href{mailto:fengjun_guo@intsig.net}{fengjun\_guo@intsig.net})
\end{itemize}

\par\vspace{1pt}

\textit{\textbf{Affiliations}}: 
\begin{itemize}[leftmargin=*, noitemsep, label={}]
    \item $^1$ Intsig Information Co., Ltd., China
\end{itemize}

\par\vspace{12pt}

\textbf{Team:} \underline{X}reflect \underline{M}aster

\textit{\textbf{Members}}:  
\begin{itemize}[leftmargin=*, noitemsep, label={}]
    \item Jiagao Hu$^1$ (\href{mailto:hujiagao@xiaomi.com}{hujiagao@xiaomi.com})  
    \item Zepeng Wang$^1$ (\href{mailto:wangzepeng5@xiaomi.com}{wangzepeng5@xiaomi.com}) 
    \item Fei Wang$^1$ (\href{mailto:wangfei11@xiaomi.com}{wangfei11@xiaomi.com}) 
    \item Daiguo Zhou$^1$ (\href{mailto:zhoudaiguo@xiaomi.com}{zhoudaiguo@xiaomi.com}) 
\end{itemize}

\par\vspace{1pt}

\textit{\textbf{Affiliations}}: 
\begin{itemize}[leftmargin=*, noitemsep, label={}]
    \item $^1$ MiLM Plus, Xiaomi Inc. 
\end{itemize}

\par\vspace{12pt}

\textbf{Team:} AIIALab

\textit{\textbf{Members}}:  
\begin{itemize}[leftmargin=*, noitemsep, label={}]
    \item Yi'ang Chen$^1$ (\href{chenya@stu.hit.edu.cn}{chenya@stu.hit.edu.cn})  
    \item Honghui Zhu$^1$ (\href{longtourister@163.com}{longtourister@163.com})
    \item Mengru Yang$^1$ (\href{ymr2200642844@163.com}{ymr2200642844@163.com})
    \item Yan Luo$^1$ (\href{luoyan1007@126.com}{luoyan1007@126.com})
    \item Kui Jiang$^1$ (\href{jiangkui@hit.edu.cn}{jiangkui@hit.edu.cn}) 
    \item Jin Guo$^1$ (\href{gjguojingj@gmail.com}{gjguojingj@gmail.com})
\end{itemize}

\par\vspace{1pt}

\textit{\textbf{Affiliations}}: 
\begin{itemize}[leftmargin=*, noitemsep, label={}]
    \item $^1$ Harbin Institute of Technology
\end{itemize}

\par\vspace{12pt}

\textbf{Team:} VIP Lab

\textit{\textbf{Members}}:
\begin{itemize}[leftmargin=*, noitemsep, label={}]
    \item Jonghyuk Park$^1$ (\href{mailto:jonghyukpark@unist.ac.kr}{jonghyukpark@unist.ac.kr})
    \item Jae-Young Sim$^1$ (\href{mailto:jysim@unist.ac.kr}{jysim@unist.ac.kr})
\end{itemize}

\par\vspace{1pt}

\textit{\textbf{Affiliations}}: 
\begin{itemize}[leftmargin=*, noitemsep, label={}]
    \item $^1$ Graduate School of Artificial Intelligence, Ulsan National Institute of Science and Technology, Republic of Korea
\end{itemize}

\par\vspace{12pt}

\textbf{Team:} YuFans

\textit{\textbf{Members}}:
\begin{itemize}[leftmargin=*, noitemsep, label={}]
    \item Wei Zhou$^1$ (\href{mailto:weichow@u.nus.edu}{weichow@u.nus.edu})
    \item Hongyu Huang$^2$ (\href{mailto:huanghy@zju.edu.cn}{huanghy@zju.edu.cn})
    \item Linfeng Li$^1$ (\href{mailto:linfeng@u.nus.edu}{linfeng@u.nus.edu})
    \item Lindong Kong$^1$ (\href{mailto:lindong@u.nus.edu}{lindong@u.nus.edu})
\end{itemize}

\par\vspace{1pt}

\textit{\textbf{Affiliations}}:
\begin{itemize}[leftmargin=*, noitemsep, label={}]
    \item $^1$ National University of Singapore
    \item $^2$ Zhejiang University, China
\end{itemize}

\par\vspace{12pt}

\textbf{Team:} KLETech-CEVI

\textit{\textbf{Members:}}

\begin{itemize}[leftmargin=*, noitemsep, label={}]
    \item Saiprasad Meesiyawar$^{4}$ (\href{mailto:saiprasad@cevi.co.in}{saiprasad@cevi.co.in})
    \item \mbox{Misbha Falak Khanpagadi$^{1,4}$ (\href{mailto:01fe23bca089@kletech.ac.in}{01fe23bca089@kletech.ac.in})}
    \item Nikhil Akalwadi$^{2,4}$ (\href{mailto:nikhil.akalwadi@kletech.ac.in}{nikhil.akalwadi@kletech.ac.in})
    \item Ramesh Ashok Tabib$^{3,4}$ (\href{mailto:ramesh_t@kletech.ac.in}{ramesh\_t@kletech.ac.in})
    \item Uma Mudenagudi$^{3,4}$ (\href{mailto:uma@kletech.ac.in}{uma@kletech.ac.in})
\end{itemize}

\par\vspace{1pt}

\textit{\textbf{Affiliations:}}

\begin{itemize}[leftmargin=*, noitemsep, label={}]
    \item KLE Technological University, Hubballi, India
    \item $^1$ Department of Computer Applications
    \item $^2$ School of Computer Science and Engineering
    \item $^3$ Department of Electronics and Communication Engineering
    \item $^4$ Center of Excellence in Visual Intelligence (CEVI)
\end{itemize}

\par\vspace{12pt}

\textbf{Team:} PSU

\textit{\textbf{Members}}: 
\begin{itemize}[leftmargin=*, noitemsep, label={}]
    \item Bilel Benjdira$^1$ (\href{mailto:bbenjdira@psu.edu.sa}{bbenjdira@psu.edu.sa}) 
    \item Anas M. Ali$^1$ (\href{mailto:aaboessa@psu.edu.sa}{aaboessa@psu.edu.sa}) 
    \item Wadii Boulila$^1$ (\href{mailto:wboulila@psu.edu.sa}{wboulila@psu.edu.sa})
\end{itemize}

\par\vspace{1pt}

\textit{\textbf{Affiliations}}: 
\begin{itemize}[leftmargin=*, noitemsep, label={}]
    \item $^1$ Robotics and Internet-of-Things Laboratory, Prince Sultan University, Riyadh 12435, Saudi Arabia
\end{itemize}

\par\vspace{12pt}

\textbf{Team:} SiGMoid

\textit{\textbf{Members}}:
\begin{itemize}[leftmargin=*, noitemsep, label={}]
    \item Kosuke Shigematsu$^1$ (\href{mailto:k-shigematsu@oita-ct.ac.jp}{k-shigematsu@oita-ct.ac.jp})
    \item Hiroto Shirono$^2$ (\href{mailto:shiro11236h@gmail.com}{shiro11236h@gmail.com})
    \item Asuka Shin$^1$ (\href{mailto:jasasu602@gmail.com}{jasasu602@gmail.com})
\end{itemize}

\par\vspace{1pt}

\textit{\textbf{Affiliations}}:
\begin{itemize}[leftmargin=*, noitemsep, label={}]
    \item $^1$ National Institute of Technology, Oita College, 1666 Maki, Oita-shi, Oita, Japan
    \item $^2$ Graduate School of Life Science and Systems Engineering, Kyushu Institute of Technology, 2-4 Hibikino, Wakamatsu-ku, Kitakyushu-shi, Fukuoka, Japan
\end{itemize}

\par\vspace{12pt}

\textbf{Team:} NTR

\textit{\textbf{Members}}:
\begin{itemize}[leftmargin=*, noitemsep, label={}]
    \item Guoyi Xu$^1$, Yaoxin Jiang$^1$, Jiajia Liu$^1$, Yaokun Shi$^1$, Jiachen Tu$^{1,*}$ (\href{mailto:jtu9@illinois.edu}{jtu9@illinois.edu})
\end{itemize}

\par\vspace{1pt}

\textit{\textbf{Affiliations}}:
\begin{itemize}[leftmargin=*, noitemsep, label={}]
    \item $^1$ University of Illinois Urbana-Champaign, USA
\end{itemize}

\par\vspace{12pt}

\textbf{Team:} refineX

\textit{\textbf{Members}}:
\begin{itemize}[leftmargin=*, noitemsep, label={}]
    \item Shreeniketh Joshi (\href{mailto:shreenikethjoshi0605@gmail.com}{shreenikethjoshi0605@gmail.com})
\end{itemize}
\par\vspace{1pt}

\textit{\textbf{Affiliations}}:
\begin{itemize}[leftmargin=*, noitemsep, label={}]
    \item KLE Technological University, India
\end{itemize}

\par\vspace{12pt}

\textbf{Team:} ACVLAB

\textit{\textbf{Members}}:  
\begin{itemize}[leftmargin=*, noitemsep, label={}]
    \item Jin-Hui Jiang$^1$ (\href{mailto:a6478492@gmail.com}{a6478492@gmail.com})  
    \item Yu-Fan Lin$^3$ (\href{mailto:aas12as12as12tw@gmail.com}{aas12as12as12tw@gmail.com}) 
    \item Yu-Jou Hsiao$^3$ (\href{mailto:re6141045@gs.ncku.edu.tw}{re6141045@gs.ncku.edu.tw})
    \item Chia-Ming Lee$^{2}$ (\href{mailto:zuw408421476@gmail.com}{zuw408421476@gmail.com})
    \item Fu-En Yang$^{4}$ (\href{mailto:fredy@nvidia.com}{fredy@nvidia.com})
    \item Yu-Chiang Frank Wang$^{4}$ (\href{mailto:frankwang@nvidia.com}{frankwang@nvidia.com})
    \item Chih-Chung Hsu$^{2,3}$ (\href{mailto:chihchung@nycu.edu.tw}{chihchung@nycu.edu.tw})
\end{itemize}

\par\vspace{1pt}

\textit{\textbf{Affiliations}}: 
\begin{itemize}[leftmargin=*, noitemsep, label={}]
    \item $^1$ Institute of Computational Intelligence, National Yang Ming Chiao Tung University
    \item $^2$ Institute of Intelligent Systems, National Yang Ming Chiao Tung University
    \item $^3$ Institute of Data Science, National Cheng Kung University
    \item $^4$ NVIDIA
\end{itemize}

\end{appendices}

%% file: Sections/Appendix-B.tex
\clearpage
\onecolumn
\begin{appendices}

\section{Extended Visual Comparisons}
\label{sec:appendix-b}

\begin{figure*}[ht]
    \centering
    \includegraphics[width=1.0\linewidth]{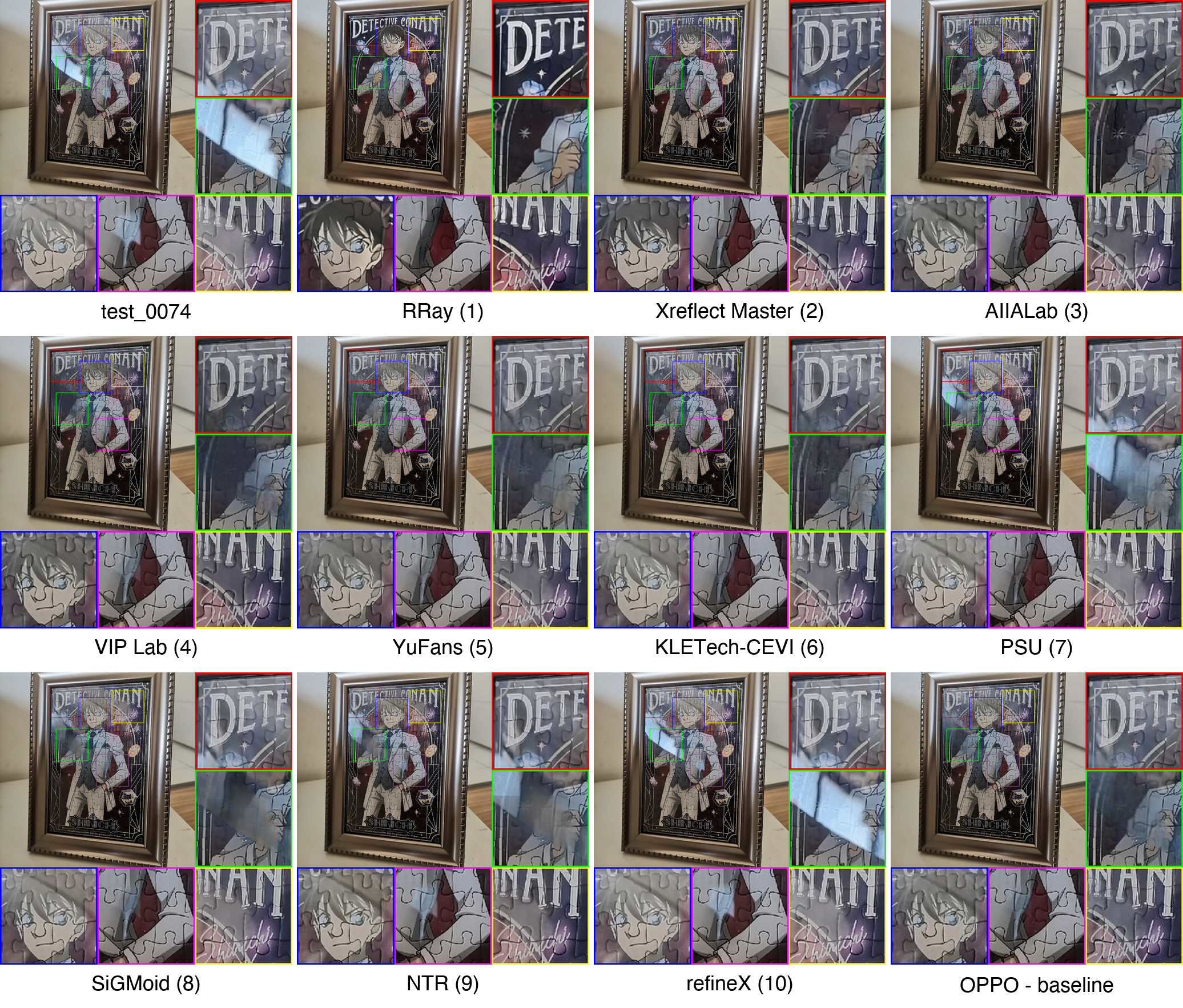}
    \caption{Visual comparison on $test_{0074}$.}
    \label{fig:vis_0074}
\end{figure*}

\begin{figure}[h]
    \centering
    \includegraphics[width=1.0\linewidth]{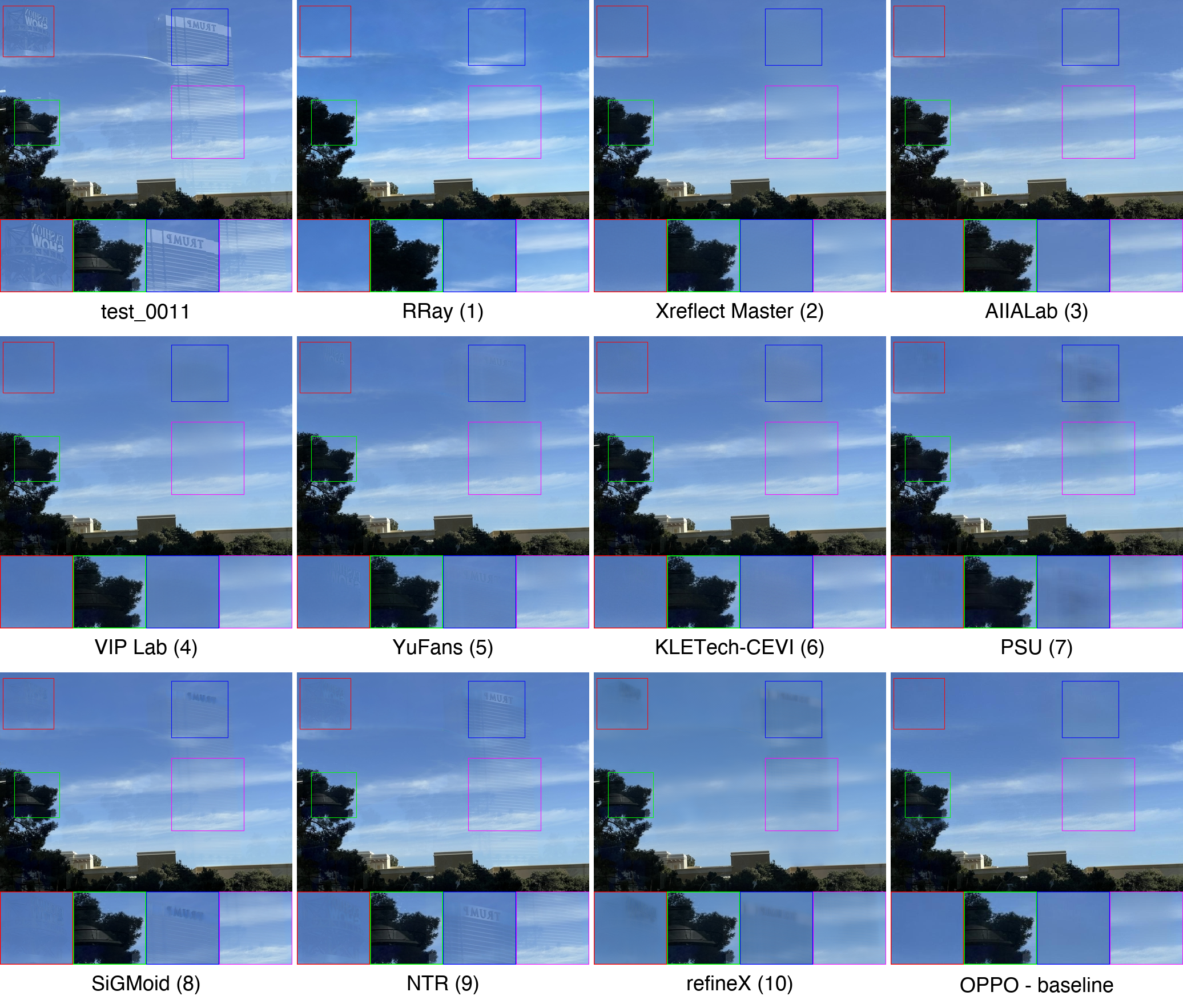}
    \caption{Visual comparison on $test_{0011}$.}
    \label{fig:vis_0011}
\end{figure}

\begin{figure}[t]
    \centering
    \includegraphics[width=1.0\linewidth]{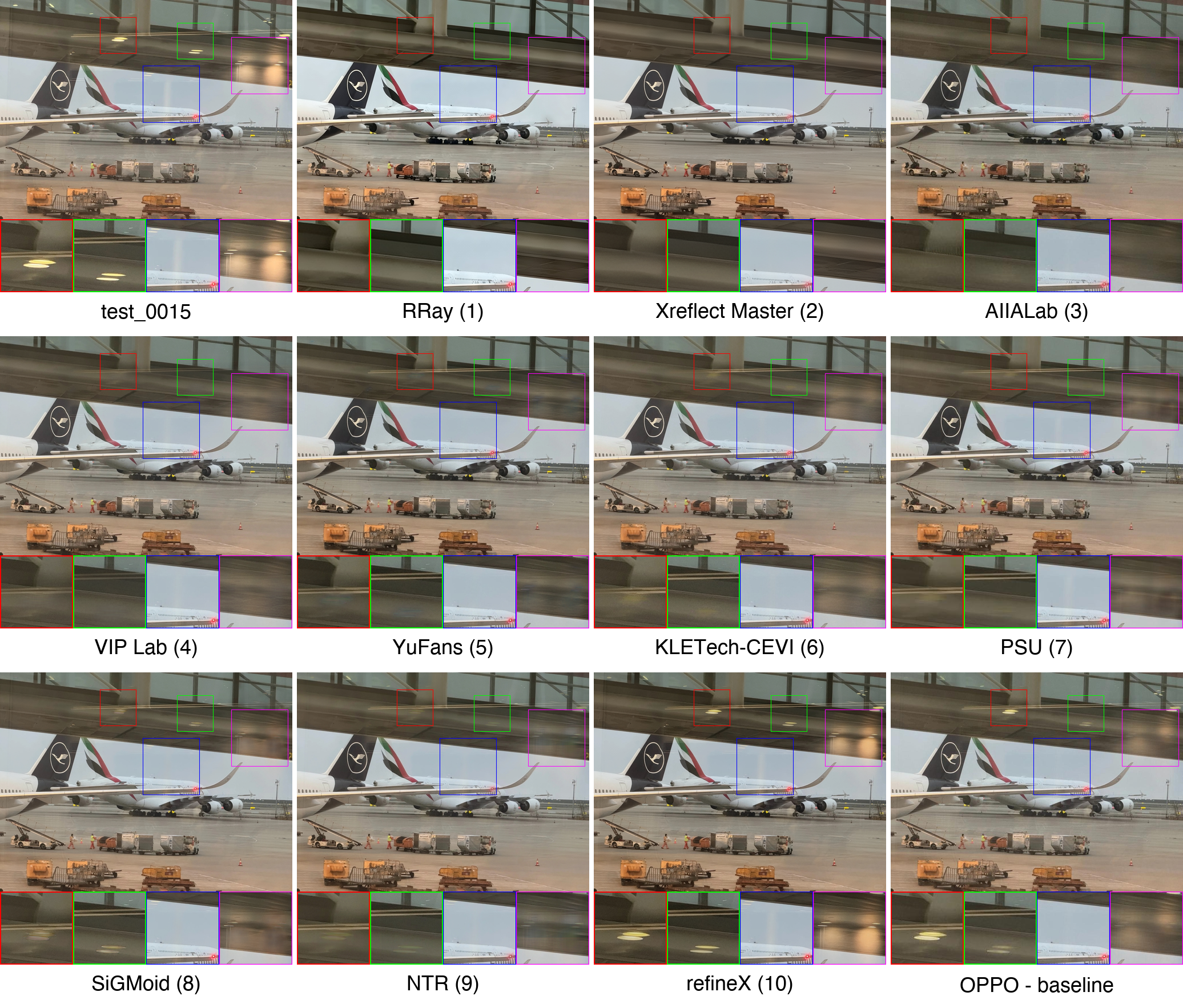}
    \caption{Visual comparison on $test_{0015}$.}
    \label{fig:vis_0015}
\end{figure}

\begin{figure}[t]
    \centering
    \includegraphics[width=1.0\linewidth]{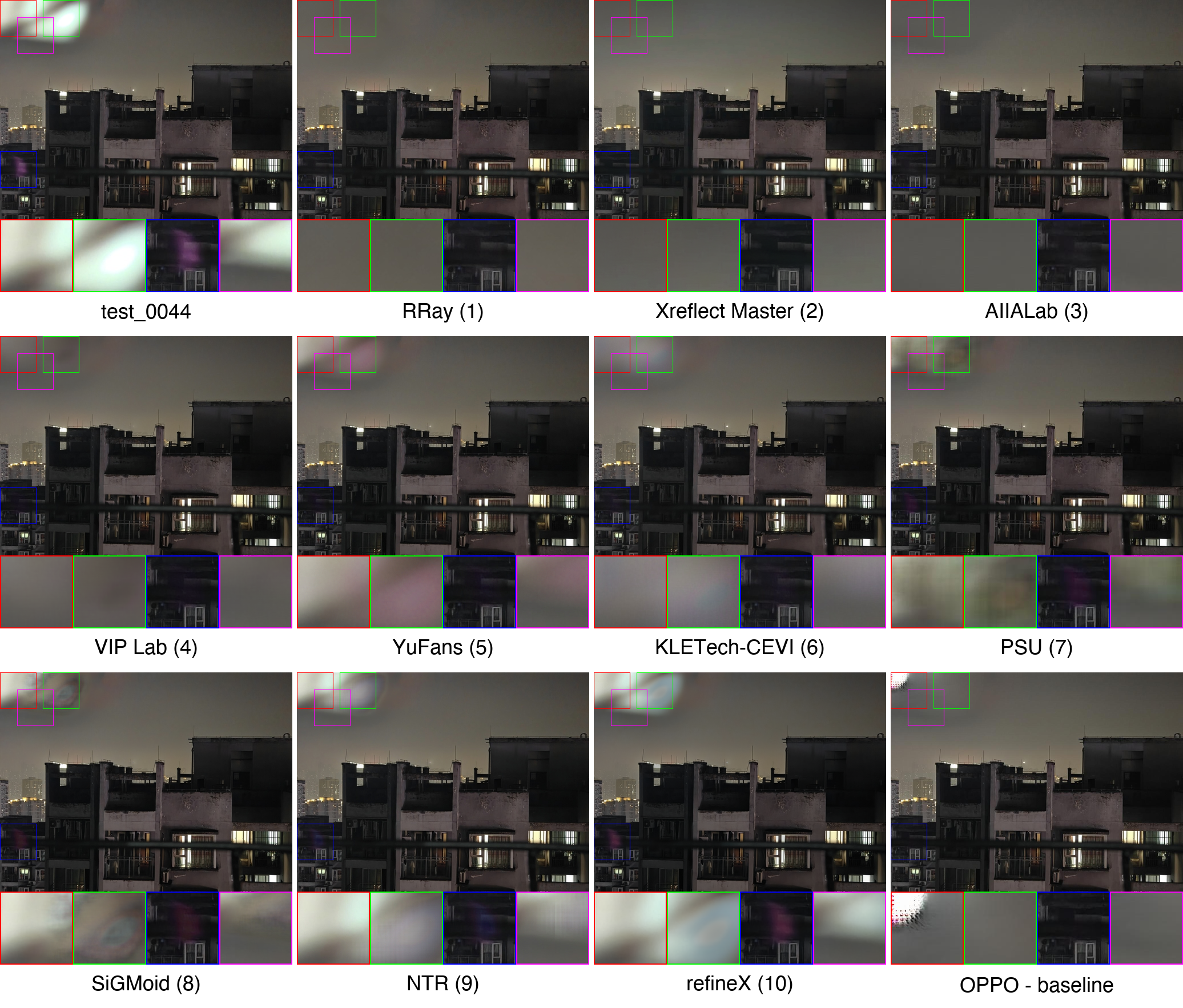}
    \caption{Visual comparison on $test_{0044}$.}
    \label{fig:vis_0044}
\end{figure}

\begin{figure}[t]
    \centering
    \includegraphics[width=1.0\linewidth]{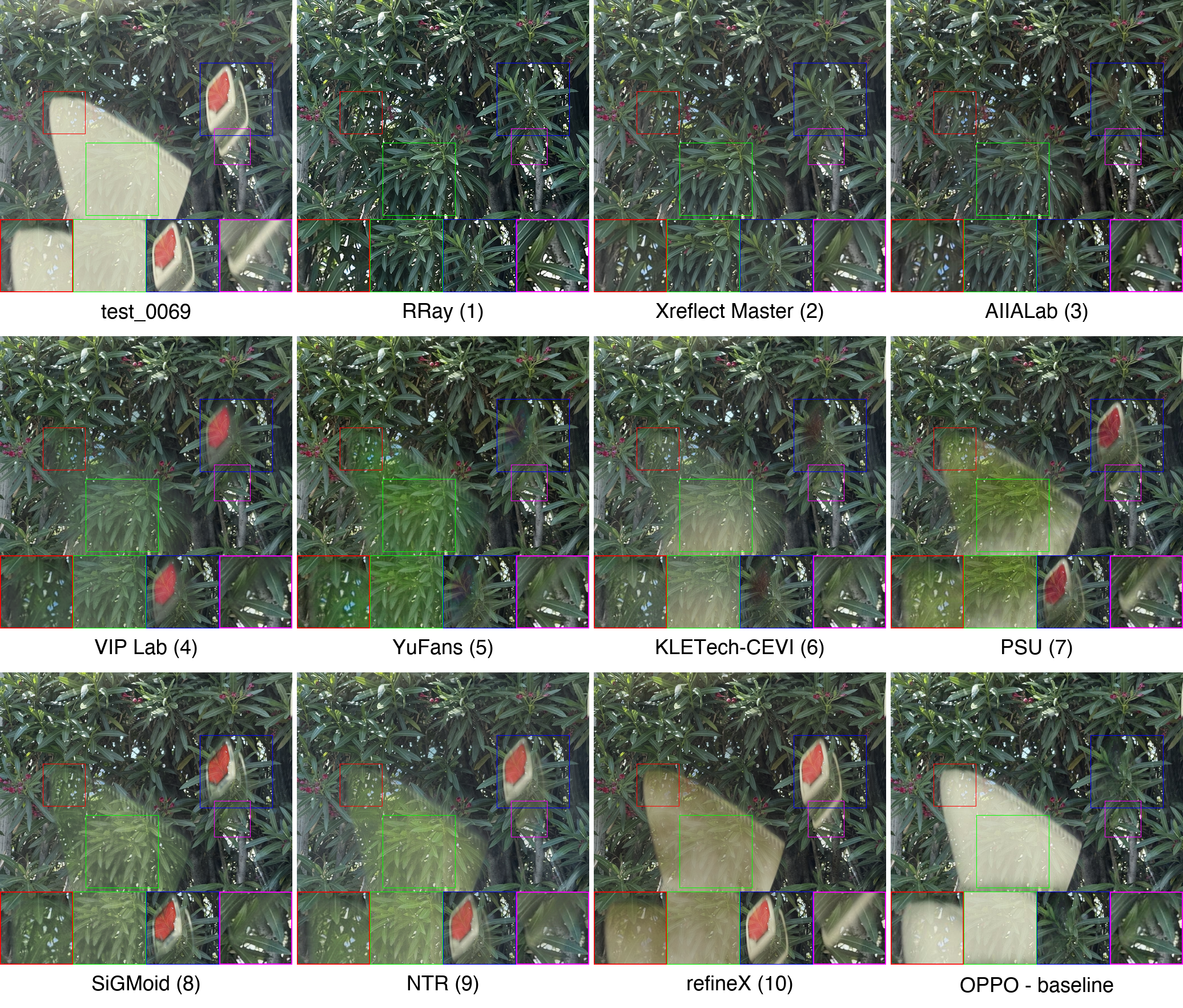}
    \caption{Visual comparison on $test_{0069}$.}
    \label{fig:vis_0069}
\end{figure}

\begin{figure}[t]
    \centering
    \includegraphics[width=1.0\linewidth]{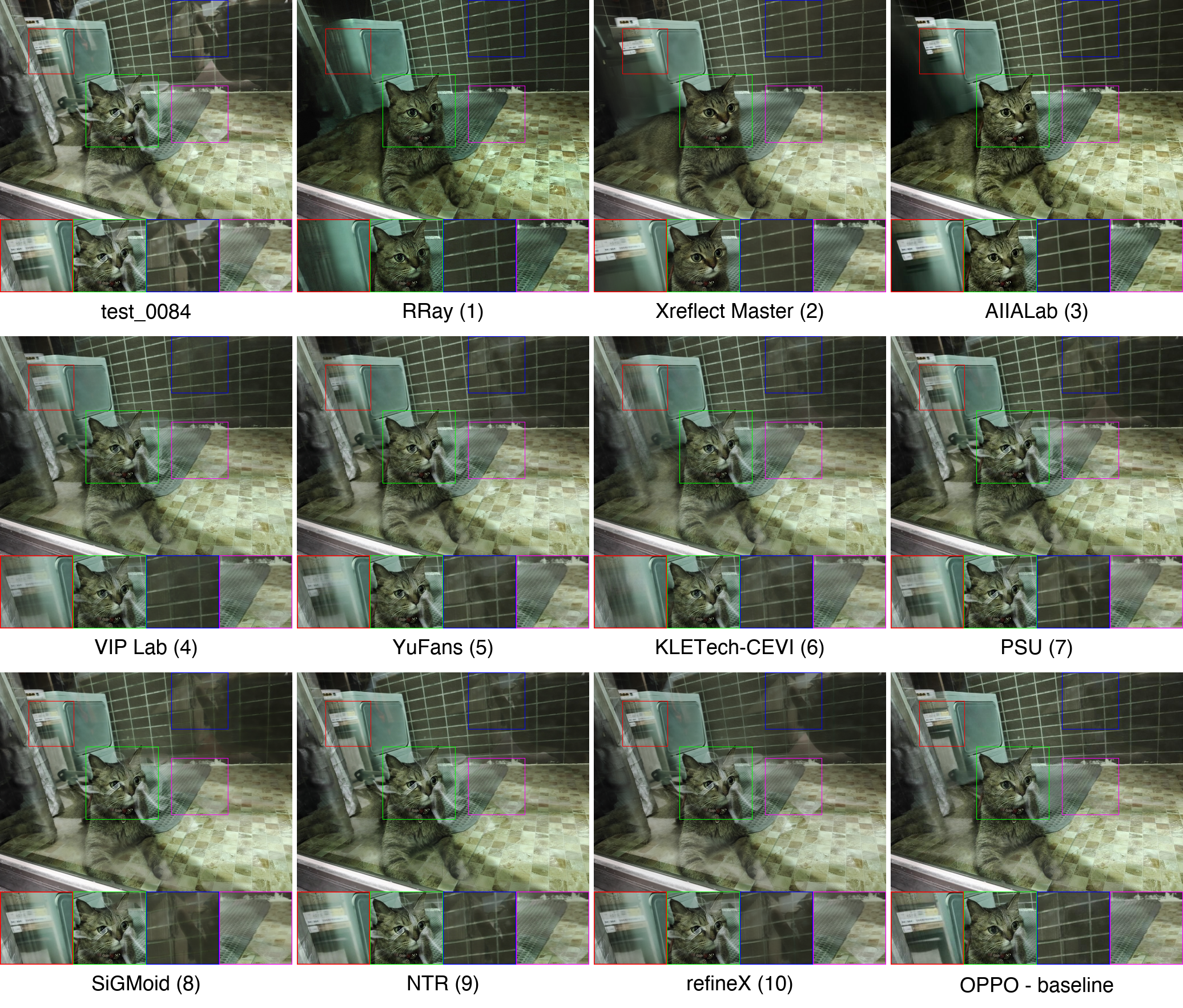}
    \caption{Visual comparison on $test_{0084}$.}
    \label{fig:vis_0084}
\end{figure}

\begin{figure}[t]
    \centering
    \includegraphics[width=1.0\linewidth]{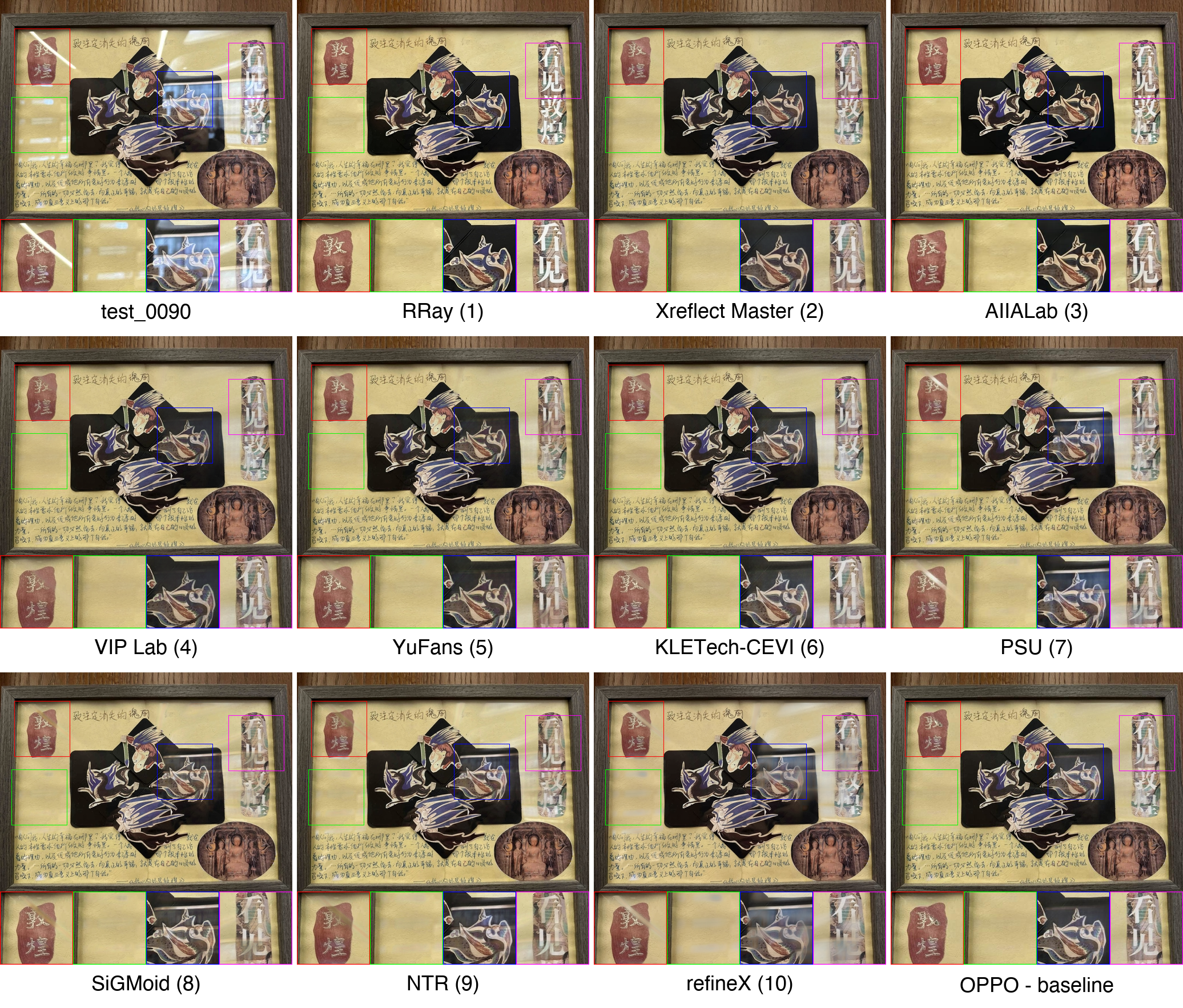}
    \caption{Visual comparison on $test_{0090}$.}
    \label{fig:vis_0090}
\end{figure}

\end{appendices}